# Human-in-the-Loop Deep Reinforcement Learning with Application to Autonomous Driving

**Authors**

Jingda Wu, Zhiyu Huang, Chao Huang, Zhongxu Hu, Peng Hang, Yang Xing, Chen Lv*,

**Affiliation**

School of Mechanical and Aerospace Engineering, Nanyang Technological University, 639798, Singapore

*Corresponding author. Email: lyuchen@ntu.edu.sg

**ABSTRACT**

Due to the limited smartness and abilities of machine intelligence, currently autonomous vehicles are still unable to handle all kinds of situations and completely replace drivers. Because humans exhibit strong robustness and adaptability in complex driving scenarios, it is of great importance to introduce humans into the training loop of artificial intelligence, leveraging human intelligence to further advance machine learning algorithms. In this study, a real-time human-guidance-based deep reinforcement learning (Hug-DRL) method is developed for policy training of autonomous driving. Leveraging a newly designed control transfer mechanism between human and automation, human is able to intervene and correct the agent's unreasonable actions in real time when necessary during the model training process. Based on this human-in-the-loop guidance mechanism, an improved actor-critic architecture with modified policy and value networks is developed. The fast convergence of the proposed Hug-DRL allows real-time human guidance actions to be fused into the agent's training loop, further improving the efficiency and performance of deep reinforcement learning. The developed method is validated by human-in-the-loop experiments with 40 subjects and compared with other state-of-the-art learning approaches. The results suggest that the proposed method can effectively enhance the training efficiency and performance of the deep reinforcement learning algorithm under human guidance, without imposing specific requirements on participants' expertise and experience.

**MAIN TEXT**

The development of autonomous vehicles (AVs) has gained increasing attention from both academia and industry in recent years (1). As a promising application domain, autonomous driving has been boosted by the ever-growing artificial intelligence (AI) technologies. From the advances made in environment perception and sensor fusion to successes achieved in human-like decision and planning, we have witnessed the great innovations developed and applied in AVs (2-4). As an alternative option to the conventional modular solution that divides the driving system into connected modules such as perception, localization, planning and control, end-to-end autonomous driving has become promising. Now it serves as a critical testbed for the development of the perception and decision-making capabilities of AI and AVs.

In terms of learning-based autonomous driving policies with an end-to-end paradigm, currently there are mainly two branches, namely, imitation learning (IL) and deep reinforcement learning (DRL). IL aims to mimic human drivers to reproduce the demonstration control actions in given states. Thanks to its intuitive and easy-to-use characteristics, IL has been applied in AV control strategies under many specific cases, including rural and urban driving scenarios (5, 6). However, two main inherent issues have been exposed in practical applications. The first one is the distributional shift, i.e. the imitation errors accumulated over time would lead to deviations from the training distribution, resulting in failures in control (9, 10). Some methods, including Dataset aggregation imitation learning (DAgger) (7), Generative adversarial imitation learning (GAIL) (8), and their derived ones, were proposed to mitigate this problem. The other issue is the limitation of asymptotic performance. Since IL's behaviour is derived from the imitation source, i.e., the experts who provide the demonstrations, therefore the performance of the learned policies is limited and could hardly surpass that of the experts. DRL, which is another data-driven self-optimization-based algorithm, shows great potential to mitigate the aforementioned issues (11-13). Constructed by exploration-exploitation and trial-and-error mechanisms, DRL algorithms are able to autonomously search for feasible control actions and optimize the policy (14). During the early stage of the DRL development, some model-free algorithms, such as Deep Q-learning (DQL) and deep deterministic policy gradient (DDPG), were popular in driving policy learning for AVs (15-17). More recently, actor-critic DRL algorithms with more complex network structures have been developed and achieved better control performances in autonomous driving (18). Particularly, the state-of-the-art algorithms, including the soft actor-critic (SAC) and twin-delayed deep deterministic policy gradient (TD3), have been successfully implemented in AVs under many challenging scenarios, such as the complex urban driving and high-speed drifting conditions (19-21).

Although many achievements have been made in DRL methods, challenges still remain. The major challenge is the sample or learning efficiency. Because in most situations, the efficiency of the interactions between the agent and environment is very low, and the model training consumes remarkable computational resources and time (22). The learning efficiency can be even worse when the reward signal generated by the environment is sparse, and thereby reward shaping methods were proposed to improve the learning efficiency in the sparse-reward environment (23, 24). Another challenge is that the DRL methods (particularly with training from scratch) exhibit limited capability of scene understanding under complex environments, which would inevitably deteriorate its learning performance and generalization capability. Therefore, in AV applications, DRL-enabled strategies are still unable to surpass and replace human drivers to handle various situations, due to their limited smartness and ability (25, 26). Besides, some emerging methods have reconsidered human characteristics and attempt to learn from common sense knowledge and neuro-symbolics (27, 28), for improving machine intelligence. As humans exhibit strong robustness and adaptability toward context understanding and knowledge-based reasoning, thus it is promising to introduce human guidance into the training loop of data-driven approaches, leveraging human intelligence to further advance learning-based methods for AVs.

Human intelligence can be reflected in several aspects in DRL training, include human assessment, human demonstration, and human intervention. Some researchers have made great efforts to introduce humans' assessments to DRL training and indeed gained success in related applications, such as simulation games (29), and robotic action control (30). However, these methods can hardly handle many other more complex application scenarios where explicit assessments are unavailable. Instead, human's direct control and guidance on agents could be more efficient for the algorithm training, and this gives rise to the architecture of incorporating DRL with learning from demonstration (LfD) (31) and learning from intervention (LfI) (32), where popular ILs, such as Vanilla-IL (33) and inverse reinforcement learning (34) are involved. Within this framework, representative algorithms were proposed based on DQL (35, 36) and DDPG (31). Some associated implementations in robotics were then reported, demonstrating improved performance comparing to original reinforcement learning (37-39). However, these methods are still far from mature. They either directly replace the output actions of DRL by using human actions or use supervised learning (SL) with human demonstrations to pre-train the DRL agent, whereas the learning algorithm architecture still remains unchanged. Recently, attempts were made on DRL structural modifications. By redefining policy functions and adding behavioral-cloning objectives, the new DRL schemes are able to effectively accelerate the training process of DRL leveraging offline human experience (40,41). Nonetheless, compared to offline human guidance, the real-time human-guidance-based schemes would be more favorable for efficient training of a DRL agent. Because for offline human-guidance-based DRLs, it is difficult to design a threshold beforehand for human intervention due to many non-quantitative factors involved. Instead, the fast scene-understanding and decision-making abilities of humans towards complex situations can be well presented via real-time human-environment interactions and further help improve the performance of DRL agents.

Nevertheless, the existing methods of DRL under real-time human guidance reported are still facing the following two main issues. Firstly, long-term supervision and guidance are exhausting for human participants (42). To adapt to the human driver's physical reactions in real-world, the procedure of the existing DRL algorithm must be slowed down in the virtual environment (43). And the extended training process would lead to a lower learning efficiency and unsatisfying human subjective feelings (36). Second, the existing DRL methods with human guidance usually require expert-level demonstrations, to ensure the quality of the data collected and achieve an ideal performance improvement. However, costly manpower and the shortage of professionals in real-world large-scale applications will limit the usage of this type of method (44). Therefore, the capability of the existing approaches, particularly the data processing efficiency, should be further improved to ensure the human-guidance-based DRL algorithms to be feasible in practice. Besides, more explorations should be conducted in order to lower the requirements of the human-guided DRL algorithms on human participants.

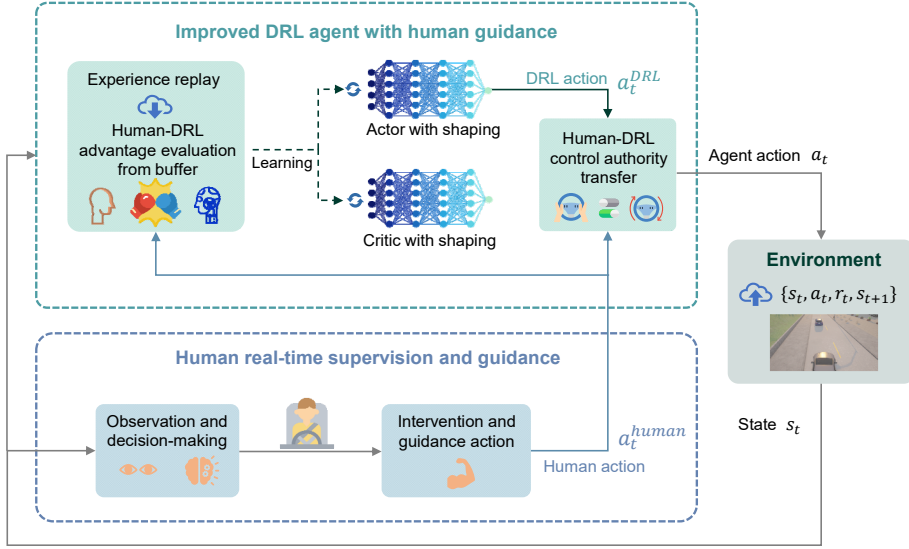

**Fig.1 The high-level architecture of the proposed Hug-DRL method with real-time human guidance.** In the proposed human-in-the-loop DRL framework, similar to the standard DRL architecture, the agent interacts with the environment during training in the autonomous driving scenario. The environment receives the output action of the agent and generates feedback, including the state transition and reward. In the meantime, the DRL agent receives and stores the state sent from the environment, keeping on optimizing its action-selection policies to improve the control performance. Beyond this, the proposed method introduces real-time human guidance into the improved DRL architecture to further enhance the agent's learning ability and performance. Specifically, the human participant observes the agent's training procedure in real-time and overrides the control of the agent by operating the steering wheel to provide guidance when he or she feels necessary. The provided human guidance action replaces the DRL policy's original action and is used as the agent's output action to interact with the environment. In the meantime, human actions are stored in the experience replay buffer. In the actor-critic algorithm, the update of the actor and critic networks is modified to be compatible with the human guidance and experience of the DRL agent. The actor network can learn from both the human guidance through imitation learning and the experience of interactions through reinforcement learning. The critic network can evaluate both the values of the agent's actions and human guidance actions. By introducing human-guided actions into both real-time manipulation and offline learning process, the training performance is expected to be significantly improved.

To bridge the abovementioned research gap and further advance the DRL method, the present work develops a human-in-the-loop DRL framework that effectively leverages human intelligence in real-time during model training. A real-time human-guidance-based DRL (Hug-DRL) method is developed and successfully applied to the agent training under autonomous driving scenarios. Under the proposed architecture, a dynamic learning process adaptively allocates weighting factors to human experience and DRL action, in order to optimize the DRL's constantly improved ability over human guidance during the overall training process. The high-level architecture of the proposed method is illustrated in Fig. 1, and the concept behind this prototype is extensively applicable beyond the specific scenario of this study. The detailed algorithms, experimental results, and the methodology adopted are reported below.

**The improved deep reinforcement learning algorithm with real-time human guidance**

In typical applications of DRL, such as the autonomous driving, the control of the DRL agent can be formulated as a Markov decision process (MDP), which is represented by a tuple $\mathcal{M}$, including the state space $\mathcal{S} \in \mathbb{R}^n$, the action space $\mathcal{A} \in \mathbb{R}^m$, the transition model $\mathcal{T}: \mathcal{S} \times \mathcal{A} \rightarrow \mathcal{S}$, and the reward function $\mathcal{R}: \mathcal{S} \times \mathcal{A} \rightarrow \mathbb{R}$, as:

$$\mathcal{M} = (\mathcal{S}, \mathcal{A}, \mathcal{T}, \mathcal{R}) \tag{1}$$

At a given step, the agent executes an action $a_t \in \mathcal{A}$ in a given state $s_t \in \mathcal{S}$ and receives a reward $r_t \sim \mathcal{R}(s_t, a_t)$. Then, the environment transitions into a new state $s_{t+1} \in \mathcal{S}$ according to the environment dynamics $\mathcal{T}(\cdot | s_t, a_t)$. In the autonomous driving scenario, the transition probability model $\mathcal{T}$ for environment dynamics is difficult to formulate. Thus, we adopt model-free reinforcement learning, in which the transition dynamics are not required to be modeled, to solve this problem.

In this work, a state-of-the-art off-policy actor-citric method, i.e., the TD3, is used to construct the high-level architecture, as shown in Supplementary Fig. 1. The TD3 algorithm chooses a deterministic action $\mu$ through the policy network $\mu$, adjusting the action-selection policy $\pi$ under the guidance of the value network $Q$. The value network approximates the value of the specific state and action based on the Bellman iterative equation. Next, TD3 sets two value networks $Q_1$ and $Q_2$ to mitigate the overestimation issue. To smooth the learning process, the target networks $\mu'$, $Q_1'$, and $Q_2'$ are adopted. The overall structure used is shown in Supplementary Fig. 2.

To realize the human-in-the-loop framework within the reinforcement learning algorithm, we combine the LfD and LfI as a uniformed architecture, where humans could decide when to intervene and override the original policy action and provide their real-time real-time actions as demonstrations. Thus, an online switch mechanism between agent exploration and human control is designed. Let $\mathcal{H}(s_t) \in \mathbb{R}^n$ denotes human's policy, and the human intervention guidance is formulated as a random event $I(s_t)$ with the observation of the human driver to the current states. Then, the agent action $a_t$ can be expressed as:

$$a_t = I(s_t) \cdot a_t^{human} + [1 - I(s_t)] \cdot a_t^{DRL} \tag{2a}$$

$$a_t^{DRL} = clip\big(\mu(s_t|\Theta^\mu) + clip(\epsilon, -c, c), a_{low}, a_{high}\big) \quad \epsilon \sim \mathcal{N}(0, \sigma) \tag{2b}$$

where $a_t^{human} \in \mathcal{H}$ is the guidance action given by a human, $a_t^{DRL}$ is the action given by the policy network, $I(s_t)$ is either equal to 0 when there was no human guidance or 1 when there is a human action occurring. $\Theta^\mu$ denotes the parameters of the policy network. $a_{low}$ and $a_{high}$ are the lower and upper bounds of the action space, $\epsilon$ is the noise subject to Gaussian distribution, and $c$ is the clipped noise boundary. The purpose of adding the Gaussian noise is to incentivize explorations in the deterministic policy. The mechanism designed by Eq. (2a) is to fully transfer the driving control authority to the human participant whenever he or she feels it is necessary to intervene in an episode during agent training.

The value network approximates the value function, which is obtained from the expectation of future reward as:

$$Q^\pi(s, a) = \mathop{\mathbb{E}}_{s \sim \mathcal{T}, a \sim \pi(\cdot|s)} \left[\sum_i^\infty \gamma^i \cdot r_i\right] \tag{3}$$

where $\gamma$ is the discount factor to evaluate the importance of future rewards, and $\mathbb{E}[\cdot]$ denotes the mathematical expectation. Let $Q(s, a)$ be the simplified representation for $Q^\pi(s, a)$, and the superscript regarding the policy would be omitted unless specified.

To solve the above expectation, the Bellman iteration is employed and the expected iterative target of the value function $y$ at the step $t$ can be calculated as:

$$y_t = r_t + \gamma \min_{i=1,2} Q'_i\left(s_{t+1}, \mu'(s_{t+1}|\Theta^{\mu'})|\Theta^{Q'_i}\right) \tag{4}$$

where $\Theta^{\mu'}$ denotes the parameters of the target policy network, and $\Theta^{Q'}$ refers to the parameters of the target value networks.

It should be noted that the two value networks with the same structure are to address the overestimation issue through clipped functionality. And it is the target policy network $\mu'$, rather than the policy network $\mu$, is used to smooth policy update. Then the loss function of the value networks in TD3 is expressed as:

$$\mathcal{L}^{Q_i}(\Theta^{Q_i}) = \mathop{\mathbb{E}}_{(s_t, a_t, r_t, s_{t+1}) \sim \mathcal{D}}[\|y_t - Q_i(s_t, a_t|\Theta^{Q_i})\|^2] \tag{5}$$

where $\Theta^{Q_i}$ denotes the parameters of the value networks, $\mathcal{D}$ denotes the experience replay buffer, which consists of the current state, action, reward, and the state of the next step.

The policy network that determines the control action is intended to optimize the value of the value network, i.e., to improve the control performance in the designated autonomous driving scenario in this study. Thus, the loss function of the policy network in TD3 algorithm is designed as:

$$\mathcal{L}^\mu(\Theta^\mu) = -\mathbb{E}[Q_1(s_t, a_t^{DRL})] = -\mathop{\mathbb{E}}_{s_t \sim \mathcal{D}}[Q_1(s_t, \mu(s_t|\Theta^\mu))] \tag{6}$$

The above formula indicates that the expectation for the policy is to maximize the value of the value network, corresponding to minimizing the loss function of the policy network. It should be noted that the unbiased estimation of the $a_t^{DRL}$ is equal to that of $\mu(s_t|\Theta^\mu)$, since the noise in Eq. (2b) is of a zero-mean distribution.

When human guidance $a_t^{human}$ occurs, the loss function of the TD3 algorithm should be revised accordingly to incorporate with human experience. Thus, the value network in Eq. (5) can be re-written as:

$$\mathcal{L}^Q(\Theta^Q) = \mathop{\mathbb{E}}_{(s_t, a_t, r_t, s_{t+1}) \sim \mathcal{D}}\left[\left(y_t - Q(s_t, a_t^{human}|\Theta^Q)\right)^2\right] \tag{7}$$

In fact, the mechanism shown in Eq. (7) modified from Eq. (4) is sufficient for establishing a human-guidance-based DRL scheme, which has been validated and reported in the existing studies (36). However, merely modifying the value network without updating the loss function of the policy network would affect the prospective performance of human guidance as stated in (40)(41). Because the value network is updated based on $\{s_t, a_t^{human}\}$, whereas the policy network still relies on $\{s_t, \mu(s_t|\Theta^\mu)\}$. This would lead to the inconsistency of the updating direction of actor and critic networks. The detailed rationale hidden behind will be analyzed in details in the Discussion section.

To address the above inconsistency issue, we modify the loss function of the policy network shown in Eq. (6) by adding a human-guidance term as:

$$\mathcal{L}^\mu(\Theta^\mu) = \mathop{\mathbb{E}}_{s_t \sim \mathcal{D}}\{-Q_1(s_t, a_t) + \omega_I \cdot [a_t - \mu(s_t|\Theta^\mu)]^2\} \tag{8}$$

where $\omega_I$ is a factor for adjusting the weightage of the human supervision loss, and the $a_t^{DRL}$ in Eq. (6) can then be replaced by simply using $a_t$ which covers both human actions and DRL policy actions. In this way, the updated direction is aligned with $\{s_t, a_t^{human}\}$ when human guidance happens. Although this generic human-guided framework has been recently proposed in some state-of-the-art methods, there are several drawbacks in their settings and thus in which further investigation and refinement are needed. For instance, the conversion between the original objective and human-guidance term is conducted rigidly, and the weighting factor of the human-guidance term is manually set and fixed (33,40,41). However, one concern is that the weighting factor $\omega_I$ is crucial for the overall learning performance of the DRL algorithm, as it determines the degree of reliance of the learning process on human guidance. Thus, it is reasonable to design an adaptive assignment mechanism for the factor $\omega_I$ that is associated with the trustworthiness of human actions. To do this, here we introduce the Temporal Difference (TD) error as an appropriate evaluation metric, and the proposed weighting factor can be modified as:

$$\omega_I = \lambda^k \cdot \left\{ \max\left[ \sup_{(s_t, a_t) \in \mathcal{D}} \exp\left(Q_1(s_t, a_t) - Q_1(s_t, \mu(s_t | \Theta^\mu))\right), 1 \right] - 1 \right\} \tag{9}$$

where $\lambda$ is a hyperparameter, which is slightly smaller than 1, and $k$ is the index of the learning episode. The temporal-decay factor $\lambda^k$ indicates that the trustworthiness of human guidance will decrease when the policy function gradually becomes matured. The clip function guarantees the policy function only learns from those "good" human guidance actions, and the exponential function amplifies the advantages brought by those "good" human guidance actions.

Intuitively, the above proposed adaptive weighting factor adjusts the trustworthiness of the human experience by quantitatively evaluating the potential results of the human's actions compared to that of the original policy. Such a mechanism forms the dynamic loss function of the policy network, instead of a fixed learning mechanism with manually tuned weighting factors reported in existing methods (33,40). Since the factor well distinguishes the varying performances of different human guidance actions provided, the requirements on the quality of human demonstration, humans' proficiency and skills, could be eased. To the best of our knowledge, it is the first time that an updating mechanism adaptive to the trustiness on human experience is proposed in the LfD/LfI-based reinforcement learning approaches, and we will show its effectiveness and advantages over the state-of-the-art techniques in the experimental validation section.

Based on Eq. (9), the batch gradient of the policy network can be given by:

$$\nabla_{\Theta^\mu} L^\mu = \frac{1}{N} \sum_t^N \{ \left( -\nabla_a Q_1(s,a)|_{s=s_t, a=\mu(s_t)} \nabla_{\Theta^\mu} \mu(s)|_{s=s_t} \right) \\ + \left( \nabla_{\Theta^\mu} (\omega_I \cdot \|a - \mu(s)\|^2 )|_{s=s_t, a=a_t} \right) \cdot I(s_t) \} \tag{10}$$

where $N$ is the batch size sample from the experience replay buffer $\mathcal{D}$.

It should be noted that, although the proposed objective function of the policy network looks similar to the control authority transfer mechanism of real-time human guidance shown in Eq. (2), the principles of these two stages, namely, the real-time human intervention and the off-policy learning, are different in the proposed method. More specifically, for real-time human intervention, the rigid control transfer illustrated by Eq. (2) enables the human's full takeover when human action occurs. For the off-policy learning, we assign weighted trustiness on human guidance without fully discarding the agent's autonomous learning, as shown in Eq. (8)-(10), allowing the learning process to be more robust.

Lastly, the originally stored tuple of the experience replay buffer is changed, and the human guidance component is then included as:

$$\mathcal{D} = \{s_t, a_t, r_t, s_{t+1}, I(s_t)\}. \tag{11}$$

In this way, the refactored DRL algorithm with real-time human guidance is obtained. The hyperparameters used and algorithm procedure are provided in Supplementary Table 1 and Supplementary Note 1, respectively.

**RESULTS**
**Overview of the experiments**
To investigate the feasibility and effectiveness of the proposed improved DRL with human guidance, a series of experiments with 40 human participants were conducted in the designed autonomous driving scenarios on a human-in-the-loop driving simulator, as shown in Fig. 2a. There were in total six typical scenarios, one of which was for the training process of the proposed method (associated with Experiments A to E), and the other five were designed for testing and evaluating the performance of the designed algorithm, as illustrated in Experiment F. The training scenario considered a challenging driving task, i.e., continuous lane-changing and overtaking, where the shaped reward of the environment encouraged non-collision and smooth driving behaviors. To successfully complete the designed tasks, in any scenario, the ego vehicle was required to start from the spawn position, keep itself within the road, avoid any collision with any other obstacles, and eventually reach the designated finishing line. As long as the ego vehicle collided with the road boundary or other traffic participants, the current episode was terminated immediately, and subsequently, a new episode would start with new spawned vehicles to continue the training process. The types, positions, and speeds of surrounding objects were varying in the testing scenarios, so as to improve the training performance of the policies under various situations with higher requirements.

To validate the training performance improvement, Experiment A was conducted by comparing the proposed method with other human-guidance-based DRL approaches. First, we implemented all related baseline DRL algorithms with the same form of real-time human guidance for convenience during the comparison. More specifically, the three baseline DRL approaches are: the IARL (with a fixed weighting factor $\omega_I$ for human guidance in the policy function of DRL (33, 40, 41), the HIRL (with shaped value function, but without modifying the policy function) (36), and the Vanilla-DRL method (the standard TD3 algorithm without human guidance) (20). All the policy networks in these are pre-initialized by supervised learning to enable faster convergence. For detailed implementations of the abovementioned approaches, please refer to the Method Section.

To investigate the effects of different human factors on the DRL training, we conducted Experiments B and C. Two key elements, i.e., human intervention mode and task proficiency, were addressed, respectively. Experiment B was conducted to explore how the different intervention modes, i.e., the continuous and the intermittent modes, as illustrated in Fig. 2c, affected the DRL training performance. The continuous mode requires more frequent human supervision and intervention, while the intermittent mode doesn't, and it allows human participates to disengage from the supervision loop for a while. The contrast was supposed to reveal the impacts of human participation frequency on the learning efficiency as well as the human subjective fatigue. It is believed that the subjects with higher proficiency or qualification regarding a specific task are usually expected to generate better demonstrations. Experiment C was designed to investigate this statement and study the correlations between human task proficiency/qualification and DRL performance improvement, as shown in Fig. 2d.

It should be noted that despite the pre-initialization, the above three experiments still started with a train-from-scratch DRL agent, denoted as "cold-start for initial training" in Fig. 2b. However, in real-world applications, such as automated driving, even if the DRL agent has been sufficiently trained beforehand, an online fine-tuning process is still needed to further improve and ensure the performance of the policy after deployment. Thus, Experiment D was designed to explore the varying effect and performance of the policies pre-trained under different algorithms throughout the fine-tuning process, as denoted by "pre-trained for fine-tuning" shown in Fig. 2b. It should be noted that the "pre-trained" here refers to the well-trained DRL policy, rather than the pre-initialization conducted by supervised-learning.

Besides, we also carried out an ablation study in experiment E to investigate the effect of pre-initialization and reward shaping on DRL's performance.

The abovementioned experiment arrangements (Experiments A-E) aim to demonstrate the superiority of the proposed method over other state-of-the-art human-guidance-based DRLs, with respect to the training efficiency and performance improvement. However, it is also necessary to test the performance of different policies in autonomous driving tasks under various scenarios. In addition, as imitation learning holds a great advantage of training efficiency due to non-interactive data generation, it would be interesting to see the performance comparisons between the IL and DRL paradigms in testing. Thus, in experiment F, we compared the driving policies obtained from the proposed Hug-DRL, the selected DRL baselines, and the IL methods (i.e., Vanilla-IL and DAgger), as illustrated in Fig. 2e. Different performance metrics under autonomous driving, including the task completion rate and vehicle dynamics states, were evaluated. Lastly, the overview of all involved experiments are provided in Fig. 2f. The statistical results are presented in the form of the mean with the standard deviation (SD). The experimental results are reported below, and the detailed methodology and experimental set-up can be found in the Methods section and supplementary files.

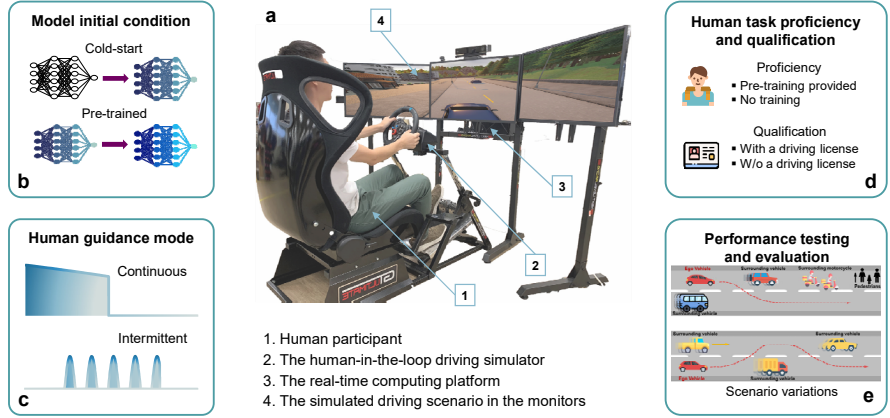

| f Method | Proficient human participant | Qualified human participant | Pre-initializing trick | Reward shaping scheme | Model initial condition | Training /Testing | |
|---|---|---|---|---|---|---|---|
| **Hug-DRL** | Both | Both | Y | 0 | Cold-start | Training | |
| **IA-RL** | Both | Both | Y | 0 | Cold-start | Training | |
| **HI-RL** | Both | Both | Y | 0 | Cold-start | Training | |
| **Vanilla-DRL** | N/A | N/A | Y | 0 | Cold-start | Training | **Experiment A** |
| **Hug-DRL** | Y | Y | Y | 1 | Cold-start | Training | |
|  | N | Y | Y | 1 | Cold-start | Training | **Experiment B** |
|  | Y | Y | Y | 1 | Cold-start | Training | |
|  | Y | N | Y | 1 | Cold-start | Training | **Experiment C** |
| **Hug-DRL** | Y | Y | N/A | 0 | Pre-trained | Training | |
| **IA-RL** | Y | Y | N/A | 0 | Pre-trained | Training | |
| **HI-RL** | Y | Y | N/A | 0 | Pre-trained | Training | **Experiment D** |
| **Hug-DRL** | Y | Y | Y | 0 | Cold-start | Training | |
|  | Y | Y | N | 0 | Cold-start | Training | |
|  | Y | Y | Y | 0 | Cold-start | Training | **Experiment E** |
|  | Y | Y | Y | 1 | Cold-start | Training | |
|  | Y | Y | Y | 2 | Cold-start | Training | |
| **Hug-DRL** | N/A | N/A | N/A | N/A | N/A | Testing | |
| **IA-RL** | N/A | N/A | N/A | N/A | N/A | Testing | |
| **HI-RL** | N/A | N/A | N/A | N/A | N/A | Testing | |
| **Vanilla-DRL** | N/A | N/A | N/A | N/A | N/A | Testing | |
| **Vanilla-IL** | N/A | N/A | N/A | N/A | N/A | Testing | |
| **Dagger-IL** | N/A | N/A | N/A | N/A | N/A | Testing | **Experiment F** |

**Fig. 2 Experimental set-up. a,** The experimental platform used in this study was a human-in-the-loop driving simulator. Key components used in the platform include a steering wheel, a real-time computation platform, three monitors, and simulated driving scenarios. **b,** There were two different initial conditions of the DRL agent during training, namely the 'cold-start' and 'pretrained'. The condition of cold-start was used in the initial training of the DRL agent, and the condition of the pre-trained policy was used for evaluating the fine-tuning performance of the DRL agent. **c,** There were two different modes of the human intervention and guidance, namely the continuous and intermittent modes, that were studied in the experiments. **d,** The human task proficiency and driving qualification, were selected as two human factors studied in this work. Their impacts on the training performance of the proposed Hug-DRL method were analyzed through experiments. **e,** Various driving scenarios were designed in the experiments for testing the control performance of the autonomous driving policies obtained by different DRL methods. **f,** Illustration of the six experiments. The numbers for illustrating the reward shaping scheme are: 0 stands for no shaping, 1 and 2 stand for two different reward shaping techniques, respectively. Detailed descriptions of the reward shaping techniques are in the Method section.

**The improved training performance of the proposed Hug-DRL method**

The results shown in Fig. 3, which were obtained from Experiment A, validate the performance improvement brought by the proposed Hug-DRL method, comparing to other state-of-the-art human-guidance-based algorithms, including the Intervention-Aided DRL (IA-RL) and Human-intervention DRL (HI-RL), as well as the Vanilla DRL without human guidance (a pure TD3 algorithm). During experiments, the obtained timestep reward and duration of each episode were recorded and assessed for each participant, to evaluate the training performance throughout an entire training session under each method. Both the episodic reward and the length of the episode were evaluated, as reflected in Figs. 3a and 3b. Based on the results, the proposed Hug-DRL method was advantageous over all other baseline methods, with respect to the asymptotic reward and training efficiency. Comparing the statistical results shown in Fig. 3c, it can be found that the averaged reward obtained with the proposed method during the entire training process was the highest at (M = -0.488, SD = 0.031), followed by that obtained with the IA-RL (M = -0.550, SD = 0.022), the HI-RL method at (M = -0.586, SD = 0.018) and the Vanilla-DRL method at (M = -0.597, SD = 0.041). Additionally, their differences were statistically significant, $F(4,36)=27.99$, according to the one-way ANOVA analysis listed in Supplementary Table 2. Moreover, the length of the episode, which accurately describes the task-completion ability, was also compared for the three methods. Based on the results shown in Fig. 3d, the mean value of the proposed method was (M = 85.1, SD = 5.2), which was advantageous over that of the HI-RL method (M = 81.5, SD = 7.3), IA-RL method (M = 78.0, SD = 8.4) and the Vanilla-DRL method (M = 64.1, SD = 9.5). Their differences were also found statistically significant, $F(4,36) =14.04$, as reflected by the ANOVA analysis listed in Supplementary Table 3. In terms of the asymptotic rewards, compared to Vanilla-DRL, the performance improvements under the proposed Hug-DRL, IA-RL and HI-RL, are 31.9%, 14.2%, and 7.1%, respectively. The above results demonstrate the effectiveness of human guidance on DRL performance enhancement.

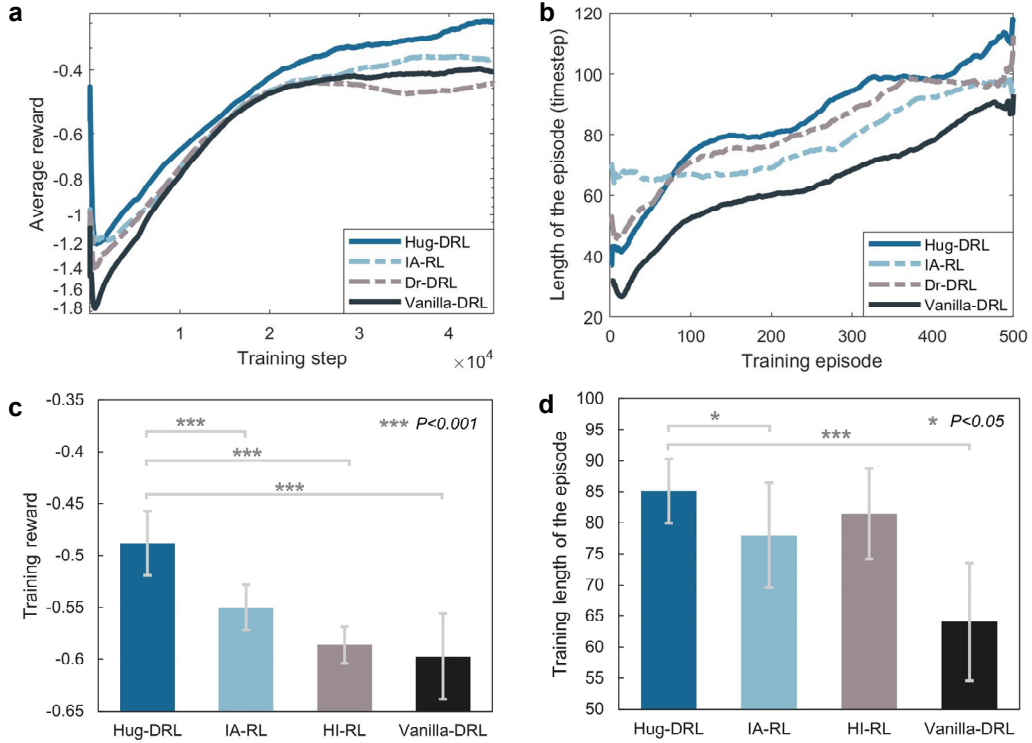

**Fig. 3 The results of the initial training performance under four different methods (Hug-DRL, IA-RL, HI-RL, and Vanilla DRL). a,** Results of the episodic training reward under different methods. The mean and SD values of the episodic training reward were calculated based on the values of the obtained rewards per episode across all subjects under each method. **b,** Results of the episodic length under the three methods. The mean and SD values of the episodic step length were calculated based on the values of the episodic length achieved per episode across all subjects under each method. **c,** Results of the average reward during an entire training session under different methods. The statistical values of the training reward were calculated based on the average value of the obtained rewards during the overall training process across all subjects under each method. **d,** Results of the average episodic length during the entire training session under different methods. The statistical values of the episodic length were calculated based on the average value of the achieved episodic length during the overall training process across all subjects under each method.

**The effects of different human guidance modes on the training performance**

We conducted two groups of testing, requiring each human subject to participate in the DRL training using the intermittent and continuous intervention modes (refer to the Methods section for detailed explanations). Example data of the episodic rewards throughout the training session for the continuous and intermittent guidance modes obtained by a representative participant was shown in Figs. 4a and 4b. From the results, it can be seen that both the continuous and the intermittent modes led to a consistently increasing trend for the episodic reward during training. Although the episodic reward increased earlier in the former case as the human intervened more frequently in the initial training phase, the final rewards achieved were on the same level for both of the two modes. The human intervention rates during the entire training sessions for the continuous and the intermittent guidance modes were further investigated, as shown in Figs. 4c and 4d. The mean values of the intervention rates (count by step) across participants for the continuous and intermittent modes were M = 25, SD = 8.3 (%) and M = 14.9, SD = 2.8 (%), respectively. Moreover, we split one training process into three separate sections, namely, the human-guided section, the non-guided section, and the overall section, and the achieved rewards were examined for each section in detail for the two intervention modes, separately. As the results illustrated in Fig. 4e, within the human intervened sections, the mean values of the training rewards for continuous and the intermittent modes were M = -0.03, SD = 0.41 and M = 0.07, SD = 0.25, respectively, but no significant statistical difference was found between the two ($p = 0.85$). Similarly, for the non-intervened sections, although the average reward of the continuous mode (M = -0.26, SD = 0.18) was better than that of the intermittent mode (M = -0.42, SD = 0.14), no significant difference could be found ($p = 0.064$). The above results indicated that in terms of the final achieved DRL performance improvement, there was no significant distinction between the continuous and intermittent modes of human guidance. However, from the angle of human workload, the intermittent mode of human guidance was found to be advantageous over the continuous mode, according to our subjective survey conducted across the participants (Supplementary Fig. 3 and Supplementary Table 4).

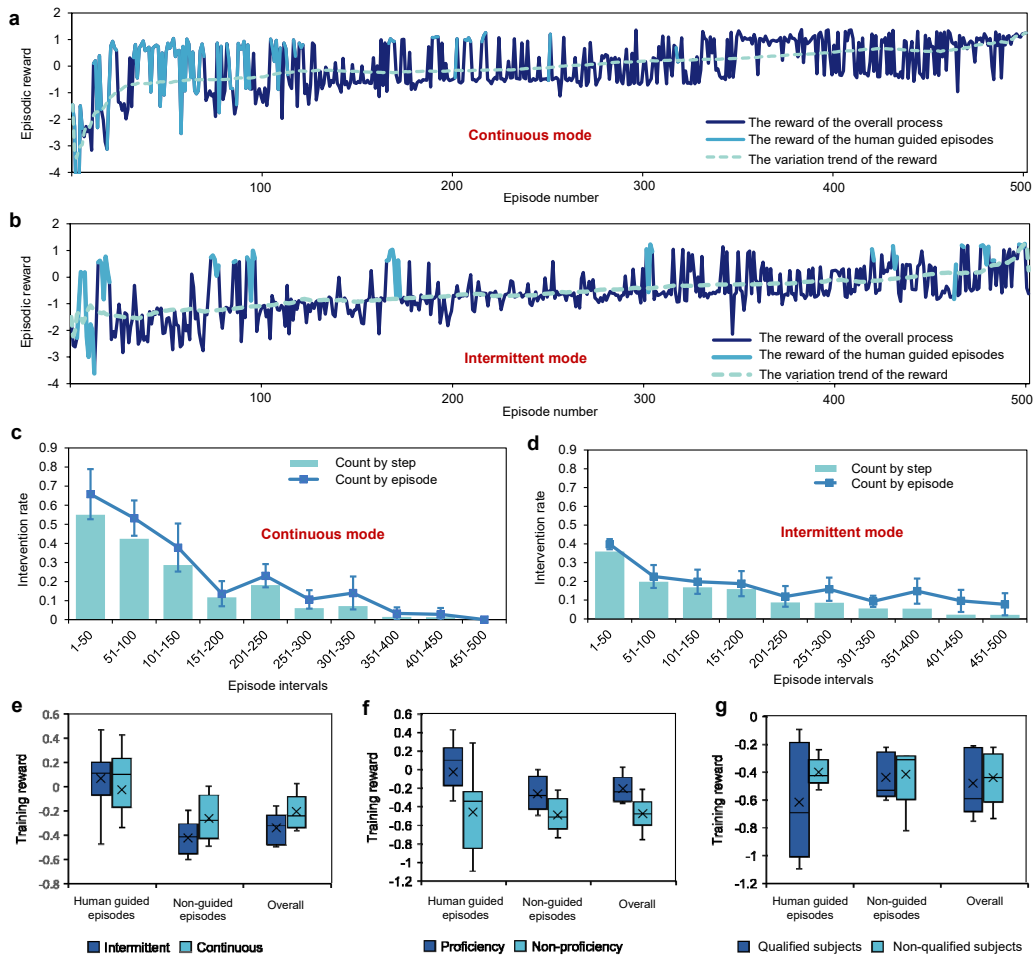

**Fig. 4 The results of the impacts of the human factors on the DRL training performance. a,** Example data of the episodic rewards over the entire training session for the continuous guidance mode obtained by a representative subject. The human-guided episodes were mainly distributed in the first half of the training process, and the guidance actions were relatively continuous. **b,** Example data of the episodic reward over the entire training session for the intermittent guidance mode obtained by a representative subject. The human-guided episodes were sparsely distributed throughout the entire training session. **c,** The human intervention rates during the entire training sessions for the continuous guidance mode. Here, two indicators, namely the one 'count by step' and the other one 'count by episode', were adopted to evaluate the human intervention rate. The former one was calculated based on the total number of steps guided by a human in a specific episodic interval, whereas the latter one was calculated based on the number of episodes intervened by a human. **d,** The human intervention rates during the entire training sessions for the intermittent guidance mode. **e,** Box plots of the training rewards achieved under the intermittent and continuous guidance modes. Under each mode, the training rewards were further analyzed based on the human-guided episodes, non-guided episodes, as well as the entire process, separately. **f,** Box plots of the training rewards achieved under the guidance provided by proficient and non-proficient participants. **g,** Box plots of the training rewards achieved under the guidance provided by qualified and unqualified participants.

**The effects of human proficiency/qualification on the training performance**

Another aspect of human factors that may have affected the DRL training performance during human guidance was task proficiency or qualification. To assuage curiosity about the correlations between the improvement of DRL performance and the task proficiency/qualification, Experiment C was conducted. As shown in Figs. 4f and 4g, the agent training rewards achieved by proficient/non-proficient and qualified/unqualified participants were illustrated and compared. Within the intervened sections, the proficient participants guided the DRL agent to gain a higher reward (M = -0.03, SD = 0.41) than non-proficient participants (M = -0.46, SD = 0.42). While for the non-intervened sections, the values of the averaged rewards guided by the proficient and non-proficient subjects were M = -0.26, SD = 0.18 and M = -0.49, SD = 0.18, separately. Considering the overall training sessions, although there was a slight difference between the two groups with respect to the training reward, i.e., M = -0.21, SD = 0.14 for the proficient group and M = -0.48, SD = 0.17 for the nonproficient one, no significant statistical difference was found between the two based on the within-group comparison ($p$=0.11). Supplementary Tables 5 and 6 provide further non-parametric ANOVA analysis of performance resulted from the standard DRL method and proficient/non-proficient participants of the proposed Hug-DRL method. In addition, no significant statistical difference was found by comparing the results obtained from the two groups of qualified and unqualified participants. According to the above comparison results, it was found that the proposed real-time human-guidance-based method had no specific requirement for the task proficiency, experience, and qualification of the participated human subjects.

**The improved online fine-tuning performance of the Hug-DRL**

As validated by the above exploration, the proposed real-time human guidance approach was capable of effectively improving the DRL performance with the initial condition of a "cold-start". Beyond this, it was very interesting to conduct Experiment D to explore the online fine-tuning ability of the proposed method, which would further improve the agent performance. As the representative examples shown in Fig. 5a, in the experiments, the participants were asked to provide guidance whenever they felt necessary within the first 10 training episodes of fine-tuning, helping the agent which originally performed at the base level to further optimize the driving policy online. Afterward, the DRL agent itself continued the rest 20 episodes until the end of the online training session. In this experiment, the proposed Hug-DRL method was compared to the other two human-guidance-based approaches, namely the IA-RL and HI-RL. Based on the resultant performance shown in Fig. 5b, within the fine-tuning stage, the proposed method and the baselines achieved similar episodic rewards (the proposed method: M= 1.02, SD = 0.36; IA-RL: M = 1.06, SD = 0.08, HI-RL: M = 1.03, SD = 0.10). However, in the subsequent session after human-guided fine-tuning, the averaged reward of the proposed method (M = 0.92, SD = 0.35) was higher than that of the IA-RL (M = 0.76, SD = 0.50), and much higher than that of the HI-RL (M = 0.19, SD = 1.01). Moreover, from the results shown in Figs. 5c and 5e, it can be found that the distribution of the episodic length obtained after fine-tuning under the proposed Hug-DRL method was more concentrated, comparing to that of the two baseline methods. The mechanism for the better performance of the Hug-DRL and IA-RL over the HI-RL after fine-tuning was also analyzed, as illustrated in Supplementary Fig. 4. In short, although the evaluation curve of the value network was updated by the human guidance action during finetuning, the policy network of the HI-RL fell into the local optima trap during the post-finetuning stage, failing to converge to the global optima (Supplementary Figs. 4a to 4c). It should be noted that Hug-DRL and IA-RL could successfully solve this issue (Supplementary Figs. 4d to 4f), and Hug-DRL achieved a better performance than IA-RL. Overall, the above results indicated that the proposed method had a great capability to fine-tune the DRL agent online, comparing to other state-of-the-art human-guidance-based DRL methods. More detailed illustrations regarding this observation can be found in the Discussion section.

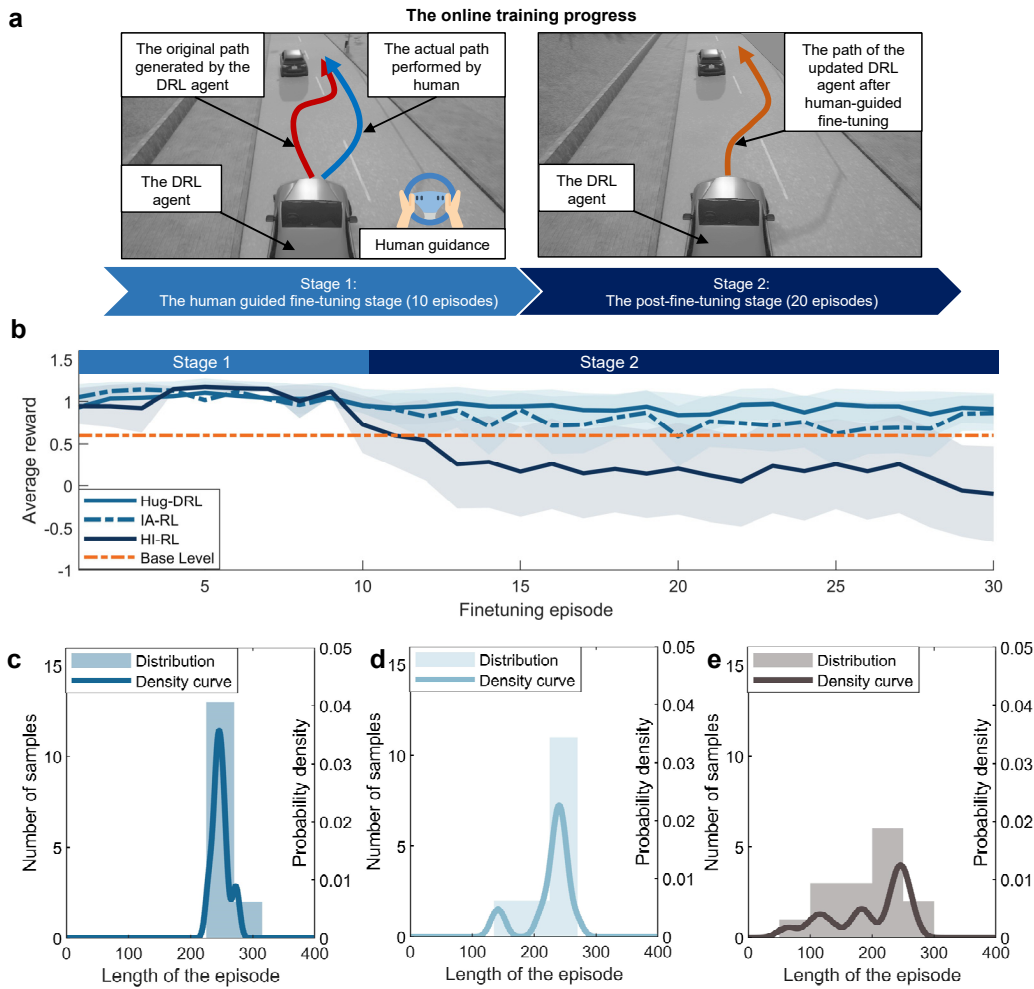

**Fig. 5 The results of the online training performance of the DRL agent under the proposed method. a,** Schematic diagram of the agent performance during the online training progress under the proposed Hug-DRL method. The entire online training progress was divided into two stages, namely stage 1: the 10-episode human-guided fine-tuning stage, and stage 2: the 20-episode non-guided post-finetuning stage. During fine-tuning, some undesirable actions of the agent were further optimized by human guidance. As a result, the performance of the DRL agent was further improved, which was reflected by the generated smooth path in the post-fine-tuning stage. **b,** The results of the episodic reward during the online training process under the proposed and two baseline approaches. Before fine-tuning, the DRL agent was pre-trained in the training scenario 0, and the average reward achieved after

the pre-training session was set as the base level for comparison in the fine-tuning stage. **c,** The distribution of the episodic length obtained under the proposed Hug-DRL method across participants during the post-fine-tuning stage. **d,** The distribution of the episodic duration obtained under the baseline IA-RL method across participants during the post-fine-tuning stage. **e,** The distribution of the episodic duration obtained under the baseline HI-RL method across participants during the post-fine-tuning stage.

**Testing of the autonomous driving policy trained by Hug-DRL under various scenarios**

To construct and optimize the configuration of the DRL-based policy, an ablation test in Experiment E was carried out for analyzing the significance of the pre-initialization and reward shaping technique. According to the results shown in Supplementary Fig. 5, we confirmed that removal of the pre-initialization process would deteriorate the training performance of the DRL agent (episodic reward: M = -0.53 for the pre-initialization scheme, M = -0.64 for the no-initialization scheme, $p$=0.005). In the meantime, it was also found that the length of the episode with mean values of 55.4, 57.2, 61.0 for non-shaping, reward-shaping scheme 1 and 2, respectively, indicating that reward shaping had no significant influence on the performance improvement.

Finally, to further validate the feasibility and effectiveness, in Experiment F, the trained model for the proposed method was tested in various autonomous driving scenarios (introduced in Supplementary Fig. 6 in detail) and compared with other five baseline methods, i.e., the IA-RL, HI-RL, Vanilla-DRL, Vanilla imitation learning (Vanilla-IL) (Supplementary Fig. 7), and the DAgger (Supplementary Fig. 8). Various testing scenarios were designed to examine the abilities of the learned policy, including the environment understanding and generalization.

The success rate of the task completion and the vehicle dynamic states (i.e., the yaw rate and the lateral acceleration) were selected as the evaluation parameters to assess the control performance of the autonomous driving agent. The heat map shown in Fig. 6a shows that the agent trained by the Hug-DRL successfully completed tasks in all untrained scenarios, while all other baselines could only complete part of the testing scenarios. Specifically, the success rate of baselines sorted by: 84.6% of the Vanilla-DRL and the DAgger, 76.9% of the HI-RL, 73.1% of the Vanilla-IL, and 65.3% of the IA-RL. In addition, the yaw rate and lateral acceleration of the agent for each method under scenario 1 were also recorded and assessed in Fig. 6b. The Hug-DRL lead to the smoothest driving behaviors with an acceleration of 0.37 $m/s^2$, and the HI-RL resulted in the most unstable driving experience (1.85 $m/s^2$). The performances of other baselines were roughly similar.

In addition to the above investigations, it was also of interest to explore the decision-making mechanism of Hug-DRL. One representative example of the testing scenarios with a trained Hug-DRL agent was shown in Figs. 6c, where the schematic diagram of the scenario, the lateral position of the ego vehicle over time, the values given the current state and action, and action of the agent were provided. As shown in Figs. 6c, approaching the two motorcycles would cause a twice decrease in $Q$ value at the current state if keeping current action, indicating a higher potential risk. Correspondingly, the ego agent would change its action to keep away from the objects and slightly drive to the left. Subsequently, the collision risk with the front bus was increased, which was reflected by the remarkably declined $Q$ value, and the DRL agent promptly decided to change lane. These results show the effects of varying surrounding traffic participants on the decision-making process of the DRL agent, and the intention and reasonable actions of the agent could be well interpreted by the results of the value evaluation function.

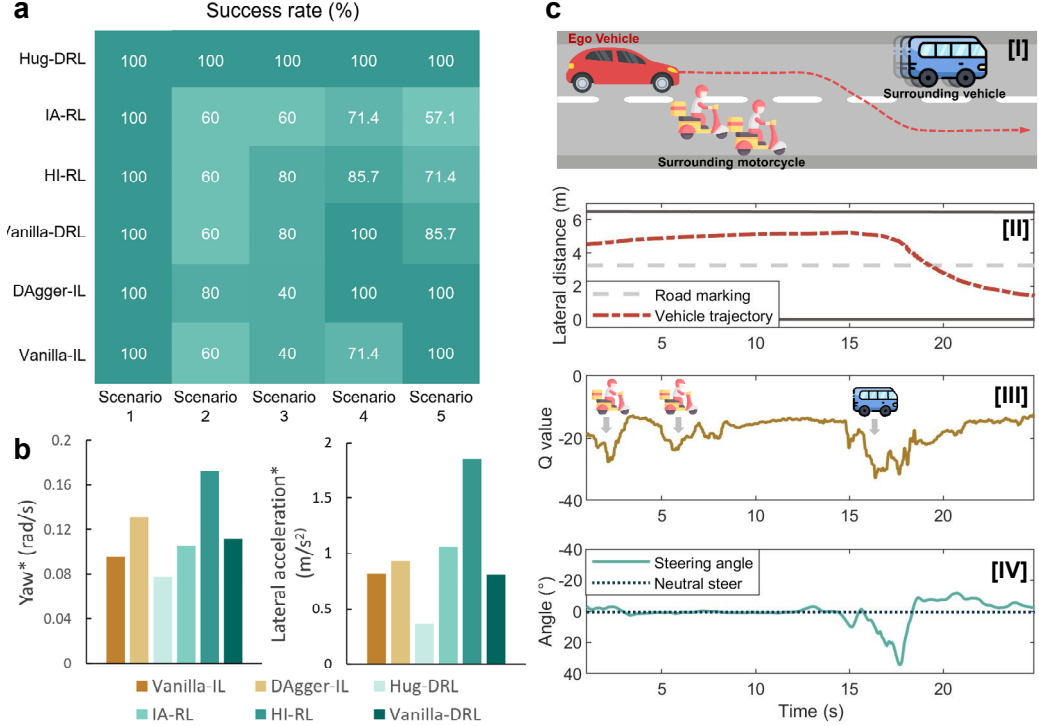

**Fig. 6 Results of the agent performance under various driving scenarios.** The agent's policy was trained by the six methods, separately. And the five scenarios, i.e., scenarios 1 to 5, were unavailable in the training process and only used for the performance testing. **a,** The success rates of the agent trained by different methods across the five testing scenarios, where the ego vehicle was spawned in different positions to calculate the success rate in one scenario. **b,** The plots of the mean of the agent's indicators under different scenarios, where two indicators: the mean of the absolute value of the yaw rate, and the mean of the absolute value of the lateral acceleration were recorded as indicators. **c,** The illustration of a representative testing scenario with an agent which was trained beforehand by using the Hug-DRL. In the testing scenario, the agent was required to surpass two motorcycles and a bus successively. Subfigures: [I] The schematic diagram of the testing scenario; [II] The lateral position of the ego vehicle; [III] The averaged Q value of the DRL agent. The value declined when the DRL agent approached the surrounding obstacles. [IV] The variation of the control action, i.e., the steering wheel angle of the DRL agent. The negative values represent left steering and the positive values correspond to the right steering actions.

## DISCUSSIONS

The existing training process of the DRL-based policy is very time-consuming and demands heavily for computing resources, especially when dealing with complex tasks with high dimensional data for scene representation. To break the above limits and to further improve the DRL algorithms by leveraging human intelligence, a novel human-in-the-loop DRL framework with human real-time guidance is proposed and investigated from different aspects in this study. Besides the proposed Hug-DRL approach, another two baseline methods with different real-time human guidance mechanisms were also implemented and compared, as well as the non-human-involved algorithms in this study. As reflected by the results shown in Fig. 3, all human-involved DRL methods were advantageous over the vanilla DRL, in terms of the training efficiency and reward achieved, demonstrating the necessity and significance of real-time human supervision and guidance at the initial training stage.

However, the reason why the introduction of real-time human guidance can effectively improve the DRL performance, should be discussed. For actor-critic DRL algorithms, the actions are determined by the policy function, of which the update is to optimize the value function, as expressed in Eq. (6). Thus, the updating rate of the policy network is constrained by the convergence rate of the value function, which relies on a relatively low-efficiency exploration mechanism. However, from the perspective of human beings who hold prior-knowledge and a better understanding of the situation with the required task, this learning is clumsy, because the agent has to experience numerous failures during the explorations before gradually converging to feasible solutions. And this forms the "cold-start" problem. However, in all human-involved DRL methods, the random and unreasonable actions are replaced by appropriate human guidance actions, which leads the value network to witness more reasonable combinations of the states and actions, effectively improving the distribution of the value function and its convergence toward the optimal point in a shorter time. Therefore, the update of the value network becomes more efficient, which further accelerates the entire training process.

Specific to the three human-involved DRL approaches, the proposed Hug-DRL approach achieves the best training efficacy and asymptotic performance, followed by IA-RL, and HI-RL performs the worst. The underlying reason is in the human guidance term of Hug-DRL and IA-RL (Eq. (8)). Specifically, in addition to the action replacement scheme of HI-RL, the human guidance term directly encourages the policy network to output human-like actions, accelerating the value function's evaluation of the acceptable policies. The subsequent problem becomes how to balance the human guidance and policy-gradient-based updating principle. The competing methods either shield gradient-term whenever human beings providing guidance or preset a fixed ratio between two terms. They fail to consider the effect of different human participants and the ever-improving ability of the DRL agent. In the proposed Hug-DRL method, the weighting assignment mechanism adaptively adjusts the dynamic trustworthiness of the DRL policy against different human guidance in the training process. In comparison to the stiff conversion mechanism of the IA-RL baseline, Hug-DRL leverages human experience more reasonably and wins higher scores, as witnessed by Fig. 3.

In addition to the above performance improvement during the training-from-scratch process, Hug-DRL was also proved to be beneficial to the online fine-tuning ability. For the learning-based approaches, including DRL, even if the models are well trained, in real-world implementations, their performance compromised due to unseen environments, unpredictable uncertainties, etc. Thus, an online fine-tuning process after deployment is of great importance for DRL applications in the real world. In this study, we evaluated the fine-tuning performance of all three methods involving human guidance, i.e., Hug-DRL, IA-RL, and HI-RL. As shown in the subplots of Figs. 5b-5e, the performance improvement of HI-RL vanished throughout fine-tuning. However, our approach successfully maintained the improved performance throughout the post-fine-tuning phase, indicating a better ability. This phenomenon could be explained by the consistency of the updates between the policy and value networks when human guidance happens. For the HI-RL model that receives human guidance, its policy network gets updated according to the objective function with $\{s, \mu(s|\Theta^\mu)\}$ in Eq. (6). however, the value network is constructed according to the $\{s, a^{human}\}$ expressed by Eq. (7). Generally, a human guidance action generates a higher true value, but it is not correctly evaluated by the value network before fine-tuning. As the online fine-tuning progress, the value network realizes the deficiency and gradually updates its output. However, sometimes the policy function can hardly catch up with the pace of the policy network's update. This means that even if the policy network has already converged close to a local optimum in the initial training phase, the change of a single point on the value function distribution that benefited from human guidance could hardly well optimize the gradient-descent-based policy function. Accordingly, the policy still updates the function around the original local optima, and thus fails to further improve itself towards the expected direction. The aforementioned inconsistency between the policy and value networks can be observed from the results shown in Supplementary Fig. 4. It should be noted that this inconsistent problem rarely occurs in the training from scratch process, due to the stronger adaptivity of the value network.

To solve the above issue, modified policy functions were proposed in Hug-DRL and IA-RL. By forcingly dragging the policy's outputs, the effect of the policy-gradient-based update was weakened in those human-guided steps, which avoids the issue of local optima trap. Thereafter, the policy could continue the noise-based exploration and gradient-based update in a space closer to the global optima. Theoretically, the inconsistency issue that occurred in HI-RL could be able to be well addressed by Hug-DRL and IA-RL. However, we found from the experimental results that IA-RL failed to achieve a competitive performance as expected, mainly due to the different forms of human guidance. Generally, the reinforcement learning agent achieves the asymptotic performance by large-scale batch training with the experience replay buffer. However, the abovementioned fine-tuning is essentially a learning process with small-scale samples. Thus, it is very difficult for IA-RL to find an appropriate learning rate in this situation, which would lead to an unstable fine-tuning performance. But the weighting factor in the proposed Hug-DRL can automatically adjust the learning rate and mitigate this issue, hence achieving the best performance as shown in Fig.5.

Besides the above discussions regarding training performance, the ability and superiority of the proposed method were also validated in testing scenarios with comparisons to other baseline approaches. More specifically, we tested the effectiveness, adaptiveness, and robustness of the proposed Hug-DRL method under various driving tasks and compared it to that of all related DRL baselines, as well as Vanilla-IL and DAgger. The results of success rate across varying test scenarios shown in Fig. 6a reflect the adaptiveness of these methods. The proposed Hug-DRL achieved the best performance across all testing scenarios among all methods. The success rates of IL approaches are significantly affected by the variations of the testing conditions, while DRL methods could well maintain their performance, indicating the better adaptiveness of DRLs. Meanwhile, the DAgger outperforms the Vanilla-IL method, achieving a compatible performance with Vanilla-DRL, but still lagging behind the proposed Hug-DRL. It is noticed that, in terms of the success rate, the performance of the IA-RL and HI-RL are worse than that of the Vanilla-DRL, which is different from the previously observed results in the training process. A feasible explanation is that some undesirable actions given by the human beings would interrupt the original training distribution of the DRL and accordingly deterioratethe robustness. Similarly, according to the results shown in Fig. 6b, the averaged yaw rate and lateral acceleration of IA-RL and HI-RL are higher than that of the Vanilla-DRL, indicating their worse performance of motion smoothness. The best performance achieved by the Hug-DRL demonstrates that, beyond the accelerated training process, the proposed human guidance mechanism can achieve effective and robust control performance during the testing process.

In addition to the above analysis, the proposed Hug-DRL method was also investigated from the aspect of human factors. The real-time human guidance has been proved to be effective for enhancing the DRL performance, however, long-term supervision may also cause unsatisfying effects, e.g., fatigue, to human participants. Fortunately, the results shown in Fig. 4e demonstrated that the intermittent guidance mode did not significantly deteriorate the performance improvement compared to the continuous mode. And the participants' subjective feelings on task workload under intermittent guidance were satisfying, according to the survey results shown in Supplementary Fig. 3. These results suggest that within the proposed human-in-the-loop DRL framework, human participants do not necessarily retain in the control loop all the time to supervise the agent training. Intermittent guidance is a good option that generates satisfying results for both agent training performance and human subjective feelings.

Moreover, we were also curious about whether the proposed Hug-DRL method had a high reliance on the participants' proficiency, skills, experience, or qualifications with respect to a specific task. As the DRL performance improvement results illustrated in Fig. 4d, the statistics indicate no significant difference between the proficient and non-proficient participant groups. This observation could also be reasonably explained by the mechanism of the proposed algorithm. Assume that a standard DRL agent is under a specific state, and the noise-based exploration can only be effective within a certain area close to the current state. Thus, the distribution is modified progressively and slowly based on the gradient update of the neural networks, being far from convergence. However, in the designed Hug-DRL method, the human guidance actions can facilitate the update of the distribution to be much more efficient. Thereafter, even if the guidance actions input from the non-proficient participants are undesirable, the explorations leveraging human guidance are still more efficient than that in the standard DRL method. Supplementary Video 1 provides a representative example of the exploration processes under the Hug-DRL and the standard DRL methods, further illustrating the above opinion in a visualized way. Similar results can be also found in Figs. 4f and 4g, where there are no significant differences between the two participant groups with and without a driving license, with respect to the achieved reward. These findings provide us more confidence that the proposed Hug-DRL method poses no high requirements on the quality of data associated with the human experience, proficiency, and task qualifications.

In summary, the above findings suggest that the proposed Hug-DRL is advantageous over the existing methods in terms of training efficiency and testing performance. It can effectively improve the agent's training performance in both the initial training and online fine-tuning stages. Intermittent human guidance can be a good option to generate satisfying results for DRL performance improvement, and at the same time it exerts no substantial burden on human workload. In particular, this new method largely reduces the requirements on the human side. The participating subjects do not need to be experts with a mastery of skilled knowledge or experience in specific areas. As long as they are able to perform normally with common sense, the DRL can be well trained and effectively improved, even if the humans' actions are undesirable. These factors make the proposed approach very promising in future real-world applications. The high-level framework, the methodology employed, and the algorithms developed in this work have great potential to be expanded to a wide range of AI and human-AI interaction applications.

**METHODS**
**Experimental platform**

The human-in-the-loop driving simulator shown in Fig. 2a was the experimental platform used for a range of experiments in this study. The technical details and the specifications of the hardware and software are reported in Supplementary Note 2 and Supplementary Table 7.

**Experimental scenarios**
In total, there were six scenarios indexed from 0 to 5 utilized in this study. The visualized scenarios are reported in Supplementary Fig. 6. The ego vehicle, i.e., the autonomous driving agent to be trained, the surrounding vehicles and pedestrians, were all spawned in a two-lane road with a width of 7 meters. Scenario 0 is only for DRL training, in which the relative velocity between the ego vehicle and the three surrounding vehicles $(v_{ego} - v_1)$ was set to +5 m/s, and the two pedestrians with random departure points at specific areas were set to cross the street. Scenarios 1 to 5 were set to evaluate the robustness and adaptiveness of learned policies under different methods. More specifically, in scenario 1, all surrounding traffic participants were removed to examine whether the obtained policies could achieve steady driving performance on the freeway. In scenario 2, we changed the positions of all the obstacle vehicles and pedestrians, and we set the relative velocity between the ego vehicle and obstacle vehicles $(v_{ego} - v_2)$ to +3 m/s, in order to generate a representative lane-change task in urban condition for the ego vehicle. In Scenario 3, the coordinates of the surrounding vehicles were further changed to form an urban lane-keeping scenario. For scenario 4, the relative velocities between the ego and the three obstacle vehicles were changed to $(v_{ego} - v_3)$=2 m/s, $(v_{ego} - v_4)$=+4 m/s, and $(v_{ego} - v_5)$= +3 m/s, respectively, and pedestrians were removed to simulate a highway driving scenario. In scenario 5, we added pedestrians with different characteristics and set various vehicle species, including motorcycles and bus, into the traffic scenario.

**Experiment design**
***The initial condition of the training***
   *The cold-start for initial training.* The initial condition of training starting from scratch was denoted as "cold-start". In this condition, the DRL agent had no prior knowledge about the environment except for the pre-initialized training.
   *The pre-trained for fine-tuning.* In this condition, the initial training with the cold-start had been completed by the agent under the standard DRL algorithm, and the agent was generally capable of executing the expected tasks. However, the behavior of the agent may still be undesirable for some situations, and thus the parameters of the algorithms were fine-tuned during this phase to further improve the agent performance.

***The human intervention activation and termination***
During the experiments, the participants were not required to intervene in the DRL training at any certain time. Instead, they were required to initiate the intervention by operating the steering wheel, providing guidance to the agent control whenever they felt it was necessary. The goal of their guidance tasks was to keep the agent on the road and try to avoid any collision, either with the road boundary or other surrounding obstacle vehicles. Once they felt that the agent was running in the correct direction with reasonable behaviors, the human participant could disengage. The detailed activation and termination mechanisms set in the experiments are explained below.
   *Intervention activation.* If a steering angle of the handwheel exceeding 5 degrees was detected, then the human intervention signal was activated, with the entire control authority transferring to the human.

*Intervention termination.* If no variation of the steering angle of the handwheel was detected for a period of time, which was set as 0.2s in this study, then the human intervention was considered to be terminated. The full control authority was transferred back to the DRL agent.

### The two human guidance modes

*The intermittent guidance.* In this mode, the participants were required to intermittently provide guidance. The entire training for a DRL agent in the designated scenario lasted for 500 episodes, and human interventions were dispersed throughout the entire training process. More specifically, the participants were only allowed to participate in 30 episodes per 100 episodes, but the determinations of whether they intervened or not, as well as when to provide guidance, depended on the participants' own decisions. For the rest of the time, the monitors were shut down to physically disengage the participants from the driving scenarios.

*The continuous guidance.* In this mode, the participants were required to continuously observe the driving scenario and provide guidance when they feel it was needed throughout the entire training session.

### Human subject proficiency and qualification

Human task proficiency was considered in this study. The proficiency of a participant was defined as follows:

*Proficient subject.* Before the experiment, the participant was first asked to naturally operate the steering wheel in the traffic scenario on the driving simulator for 30 minutes to become proficient in the experiment scenario and device operation.

*Non-proficient subject.* The participant was not asked to do the training session and instead asked to conduct the experiment directly.

Besides the proficiency, the qualification of driving was also considered in this work.

*Qualified subject.* A participant who owned a valid driving license was considered to be a qualified subject.

*Unqualified subject.* A participant who did not own a valid driving license was regarded as an unqualified subject.

### Experimental Tasks

In this work, multiple experimental tasks were designed.

*Experiment A.* The purpose of this experiment was to test the performance of the proposed Hug-DRL method and compare the performance with that of the selected baseline approaches. In total, ten participants holding a valid driving license were assigned to conduct this experiment. Before the experiment, the participants were asked to do a 30-min training session on the driving simulator to become proficient in the experiment scenario and in device operation. During the experiment, each participant was asked to provide intermittent guidance for the proposed Hug-DRL method and baseline methods, i.e., the IA-RL and HI-RL. However, the participants were not informed about the different algorithms used in the tests. In addition, the Vanilla-DRL method was used to conduct the agent training 10 times without human guidance. The initial condition of the training was set as cold-start, and the driving scenario of this experiment was set as the above-mentioned scenario 0. In addition, each participant for this experiment was required to complete a questionnaire after their tests to gather his or her personal subjective opinion on the workload level. The workload was rated on a scale from one (very low) to five (very high).

*Experiment B.* The purpose of this experiment was to study the impact of the human guidance modes on the agent performance improvement for the proposed Hug-DRL method. The same ten participants recruited in Experiment A were assigned to conduct this experiment. Before the experiment, the participants were asked to do a 30-min training session on the driving simulator to become proficient in the experiment scenario and in device operation. During the experiment, each participant was asked to provide guidance to the driving agent in continuous guidance mode for the proposed Hug-DRL method. The initial condition of the training was set as cold-start, and the driving scenario of this experiment was set as the above-mentioned scenario 0. In addition, each participant of this experiment was required to complete a questionnaire after their tests to gather his or her personal subjective opinion on the workload level. The workload was rated on a scale from one (very low) to five (very high).

*Experiment C.* The purpose of this experiment was to study the impact of human proficiency and driving qualification on the performance improvement for the proposed Hug-DRL method. Ten new subjects were recruited to participate in this experiment. Among them, five subjects holding valid driving licenses were considered as qualified participants, and the other participants without a driving license were seen as unqualified participants. Before the experiment, the participants were not provided with a training session. Instead, they were asked to directly participate in the agent training experiment. During the experiment, each participant was asked to provide continuous guidance to the driving agent for the proposed Hug-DRL method. The initial condition of the training was set as cold-start, and the driving scenario of this experiment was set as the above-mentioned scenario 0.

*Experiment D.* The purpose of this experiment was to study the online fine-tuning ability of the proposed Hug-DRL method and compare the fine-tuning ability to that of the selected baseline methods. In this experiment, the initial condition of the training was set as fine-tuning, rather than cold-start. Fifteen new participants were recruited to conduct this experiment. Before the experiment, the participants were provided with a short training session to become used to the environment and the devices. The entire fine-tuning phase was set to 30 episodes in total. During the experiment, the subjects were only allowed to intervene in the agent training within the first 10 episodes, providing guidance when needed. For the next 20 episodes, the participants were disengaged from the tasks. However, the agent's actions were continually recorded to assess its performance. Each participant was asked to conduct this experiment under the proposed Hug-DRL and the baseline methods, i.e., the IA-RL and HI-RL. Before experiments, the participants were not informed about the different algorithms used in the tests. The driving scenario of this experiment was set to scenario 0.

*Experiment E.* The purpose of this experiment was to test the impacts of the adopted pre-initialized training and the reward-shaping techniques on the training performance. In the ablation group 1, five participants, were required to complete the task in Experiment A, and the used Hug-DRL agent was not pre-trained by supervised learning. The results were obtained to compare with that of pre-trained Hug-DRL obtained in the training process. A similar set-up was conducted in the ablation group 2, and the adopted Hug-DRL agents were equipped with three different types of reward schemes: no reward shaping, reward shaping route 1, and reward shaping route 2. And in each subgroup experiment, 5 participants were asked to complete the task of Experiment A. The details of different reward shaping schemes are explained later in Eq. (21) and Eq. (22).

*Experiment F.* The purpose of this experiment was to test and compare the performance of the autonomous driving agent trained by different methods under various scenarios. We first completed the training process of two imitation learning-based policies, i.e., Vanilla-IL and DAgger. Human participants were asked to operate the steering wheel, controlling the IL agent to complete the same overtaking maneuvers as that of DRL agents (collision avoidance with surrounding traffic participants). For the Vanilla-IL, the agent was fully controlled by human participants, and there was no agent to interact with humans through control authority transfer. Gaussian noise was injected into the agent's actions for the purpose of data augmentation. The collected data was used for offline supervised learning to imitate human driving behaviors. For the DAgger, the agent learns to improve the control capability lfrom human guidance. In one episode, whenever a human participant felt the necessity of intervention, they obtained a partial control authority, and only their guidance actions would be recorded to train the DAgger agent in real-time. Since the agent was refined through the training episodes, DAgger was expected to collect more data and obtain a more robust policy than the Vanilla-IL. The tested methods included the Hug-DRL, IA-RL, HI-RL, Vanilla-DRL, DAgger and Vanilla-IL. The driving scenarios used in this experiment included the designed scenarios 1-5.

**Baseline Algorithms**

*Baseline A: Intervention-Aided DRL (IA-RL).* In this method, human guidance was introduced into the agent training process. The human actions directly replaced the output actions of the DRL, and the loss function of the policy network was also modified to fully adapt to human actions when guidance occurs. This method was derived and named from existing work reported in [29] and [32], and was further modified in this work to adapt to the off-policy actor-critic DRL algorithms. The detailed algorithm for this approach can be found in Supplementary Note 3, and the hyperparameters are listed in Supplementary Table 1 and Supplementary Table 8.

*Baseline B: Human-intervention DRL (HI-RL).* In this method, human guidance was introduced into the agent training process, however, the human actions were used to directly replace the output actions of the DRL agent without modifying the architecture of the neural networks. As a result, human actions only affect the update of the value network. This baseline approach, which was derived and named from the work reported in (36), was further modified to adapt the actor-critic DRL algorithm in our work. The detailed algorithm can be found in Supplementary Note 4, and the hyperparameters are listed in Supplementary Table 1 and Supplementary Table 8.

*Baseline C: Vanilla-DRL.* A standard DRL method (the TD3 algorithm). It was used as a baseline approach in this work. The detailed algorithm can be found in Supplementary Note 5, and the hyperparameters are listed in Supplementary Table 1 and Supplementary Table 8.

*Baseline D: Vanilla Imitation Learning (Vanilla-IL).* The Vanilla imitation learning with data augmentation was also adopted as one of the baseline methods. In this study, a deep neural network with the Vanilla-IL method was used to develop the autonomous driving policy for comparison with other DRL-based approaches. The detailed mechanism of this method is introduced in Supplementary Fig. 7. The detailed procedures of data collection and model training under Vanilla-IL are introduced in Supplementary Note 6. The hyperparameters are listed in Supplementary Table 9, and the network architecture is illustrated in Supplementary Table 8.

*Baseline E*: *Dataset Aggregation imitation learning* (*DAgger*). An imitation learning method with real-time human guidance. Under this approach, the human participants serve as experts to supervise and provide necessary guidance to an actor agent, which would learn from human demonstrations and improve its performance through training. The detailed mechanism of DAgger is illustrated in Supplementary Fig. 8. The detailed procedures of data collection and model training are introduced in Supplementary Note 6. The hyperparameters are listed in Supplementary Table 10, and the network architecture is illustrated in Supplementary Table 8.

**Implementation of the proposed Hug-DRL method in Autonomous Driving**

The proposed Hug-DRL method is developed based on TD3 with the introduction of real-time human guidance. For the DRL algorithm, appropriate selections of the state and action space, as well as the elaborated reward function design, are significant for efficient model training and performance achievement. In this work, the target tasks for the autonomous driving agent are set to complete lane-change and overtaking under designed various scenarios. To better demonstrate the feasibility, effectiveness and superiority, the challenging end-to-end paradigm is selected as the autonomous driving configuration for proof-of-concept of the proposed method. Specifically, non-omniscient state information is provided to the policy, and the state representation is selected for semantic images of the driving scene through a single channel representing the category of 45×80 pixels:

$$s_t = \{p_{ij,t} | p \in [0,1]\}_{45 \times 80}, \qquad (12)$$

where $p_{ij}$ is the channel value of the pixel $i \times j$ normalized into [0,1]. The semantic images were obtained from the sensing information provided by the simulator.

The steering angle of the handwheel is selected as the one-dimensional action variable, and the action space can be expressed as:

$$a_t = \{\alpha_t | \alpha \in [0,1]\} \qquad (13)$$

where $\alpha$ is the steering wheel angle normalized into [0,1], where the range [0,0.5) denotes the left-turn command and (0.5,1] denoted the right-turn command. The extreme rotation angle of the steering wheel is set to ±135 degrees.

The reward function should consider the requirements of real-world vehicle applications, including driving safety and smoothness. The basic reward function is designed as a weighted sum of the metrics, which is given by:

$$r_t = \tau_1 c_{side,t} + \tau_2 c_{front,t} + \tau_3 c_{smo,t} + \tau_4 c_{fail,t} \qquad (14)$$

where $c_{side,t}$, denotes the cost for avoiding a collision with the roadside boundary, $c_{front,t}$ is the cost for collision avoidance with a front obstacle vehicle, $c_{smo,t}$ is the cost of maintaining vehicle smoothness, $c_{fail,t}$ is the cost of a failure causing the termination of the episode. $\tau_1$ to $\tau_4$ are the weights of each metric.

The cost of the roadside collision is defined by a two-norm expression as:

$$c_{side,t} = -\|1 - f_{sig}(\min[d_{left,t}, d_{right,t}])\|_2 \tag{15}$$

where $d_{left}$ and $d_{right}$ are the distance to the left and right roadside boundaries, respectively. $f_{sig}$ is the sigmoid-like normalization function transforming the physical value into [0,1].

The cost for the front obstacle avoidance is defined by a two-norm expression as:

$$c_{front,t} = \begin{cases} -\|1 - f_{sig}(d_{front})\|_2 & \text{, if a front obstacle exists} \\ 0 & \text{, otherwise} \end{cases} \tag{16}$$

where $d_{front}$ is the distance to the front obstacle vehicle in the current lane.

The cost of maintaining smoothness is designed as:

$$c_{smo,t} = -\left(\frac{d\alpha_t}{dt} + (\alpha_t - 0.5)\right) \tag{17}$$

The cost of failure can be expressed as:

$$c_{fail,t} = \begin{cases} -1 & \text{if fail} \\ 0 & \text{otherwise} \end{cases} \tag{18}$$

The above reward signals stipulate the practical constraints, yet the feedback is still sparse and it does not boost the exploration behaviors, which means the DRL could easily get trapped into the local optima. The reward shaping technique is an effective tool to prevent such issue. The reward shaping transforms the original rewards by constructing an additional function, with the aim of encouraging explorations while maintaining the learned policy invariance. In this study, we compare two different reward shaping methods and a vanilla reward form to conduct the ablation study in Experiment E.

The reward shaping method 1 refers to a potential-based reward shaping function, which is well-known for its straightforward and efficient implementation (45). A typical potential-based reward shaping function $\mathcal{F}: \mathcal{S} \times \mathcal{A} \times \mathcal{S} \to \mathbb{R}$ can be written as:

$$\mathcal{F}(s_t, a_t, s_{t+1}) = \gamma\phi(s_{t+1}) - \phi(s_t) \quad \forall s_t \in \mathcal{S} \tag{19}$$

where $\phi: \mathcal{S} \to \mathcal{A}$ is a value function, which ideally should be equal to $\mathbb{E}_{a \sim \pi(\cdot|s)}[Q(s, a)]$. Since the accurate values of $Q$ are intractable before the training convergence, prior knowledge regarding the task requirement becomes a heuristic function as $\phi$ to incent the DRL's exploration. Accordingly, the function $\mathcal{F}^1$ in the adopted method 1 is defined to be associated with the longitudinal distance from the spawn point, which can be calculated as:

$$\mathcal{F}_t^1 = P_{y,t}(s_t, a_t) - P_{y,spawn} \tag{20}$$

where $P_{y,t}$ and $P_{y,spawn}$ are the current and the initial positions of the agent in the longitudinal direction, respectively. It indicates that the agent is encouraged to move forward and explore further, keeping itself away from the spawn position.

The reward shaping method 2 refers to a state-of-the-art technique named NGU (24). Its main idea is also to encourage explorations and resist frequent visits of those state values seen before.

$$\mathcal{F}_t^2 = r_t^{episode} \cdot \min\left\{\max\left\{1 + \frac{\|f(s_{t+1}|\psi) - f(s_{t+1})\| - \mathbb{E}[f(s_{t+1}|\psi)]}{\sigma[f(s_{t+1}|\psi)]}, 1\right\}, L\right\} \tag{21}$$

where $f(\cdot|\psi)$ and $f(\cdot)$ are embedded neural networks with fixed weights $\psi$ and adjustable weights, respectively. The norm $\|\cdot\|$ is to calculate the similarity between the embedded state feature, $\sigma$ denotes the standard deviation operation, $L$ is a regularization hyperparameter. The overall idea of employing $f(\cdot)$ is to assign higher additional rewards into the unvisited states, particularly during the training process (refer to (24) for details). $r_t^{episode}$ is also to encourage to explore at unvisited states, particularly during the current episode. The utilized hyperparameters are provided in Supplementary Table 8.

Thus, the overall reward function can be obtained by adding the terms $\mathcal{F}_t^1$ and $\mathcal{F}_t^2$ to the original function $r_t$.

Finally, the termination of an episode with successful completion of the task is defined as surpassing the last obstacle vehicle and reaching the finishing line without any collisions. With the above steps, the detailed implementation of the standard DRL in the designed driving scenario is completed.

For the proposed Hug-DRL, real-time human guidance is achieved by operating the steering wheel in the experiments. Thus, the steering angle of the handwheel is used as the human intervention signal, and a threshold filtering unexpected disturbance is required. Here, the event of human intervention and guidance is marked as:

$$I(s_t) = \begin{cases} 1, if \ (\frac{d\alpha_t}{dt} > \varepsilon_1) \cap \neg q \\ 0, \quad otherwise \end{cases} \quad (22)$$

where $\varepsilon_1$ is the threshold, set as 0.02. $q$ denotes the detection mechanism of human intervention termination, which is defined as:

$$q = \prod_{t}^{t+t_N} \left(\frac{d\alpha_t}{dt} < \varepsilon_2\right) \quad (23)$$

where $\varepsilon_2$ is the threshold set to 0.01, $t_N$ is the time threshold for determining the intervention termination and it is set to 0.2s as mentioned above.

For the proposed Hug-DRL method, when human participants engage in or disengage from the training process, the control authority of the agent is transferred between the human and the DRL algorithm in real-time. The detailed mechanism of control transfer is illustrated in Eq. (2).

**Participants**

In total, 40 participants (26 males, 14 females) in the age range of 34 to 21 (M = 27.43, SD= 3.02) were recruited for the experiments. The study protocol and consent form were approved by the Nanyang Technological University Institutional Review Board, protocol number IRB-2018-11-025. All research was performed per relevant guidelines/regulations. We also confirm that informed consent was obtained from all participants. All of the participants had no previous knowledge of the research topic and they had never previously experienced real-time intervention and guidance during model training in the driving scenario. Before the experiments, the participants were informed that the DRL agent would receive their guidance and improve its performance as the training process went on.

**Statistical analysis**

*Statistical methods.* A statistical analysis of the experimental data was conducted for the designed experiments. The statistical analysis was performed in Matlab (R2020a, MathWorks) using the Statistics and Machine Learning Toolbox and in Microsoft Excel. For the data shown in Figs. 3, 5, and 6, since these data generally obeyed the normal distribution, the difference of the mean values between two groups was determined using paired *t*-tests (with the threshold level α = 0.05), and the difference for multiple groups was determined using one-way ANOVA analysis. To investigate the statistical significance of the difference between data groups, as shown in Fig. 4, the non-parametric tests, including the Mann-Whitney U-test and the Kruskal-Wallis test, were adopted at the α = 0.05 threshold level.

*Definition of the evaluation metrics.* The following metrics were adopted in this study to evaluate the agent performance. The reward, reflecting the agent's performance, was chosen as the first metric. For both the step reward and episodic reward, the mean and SD values were calculated and used when evaluating and comparing the agent performance across different methods and different conditions throughout the paper. The length of the episode, which could be obtained by calculating the number of steps in one episode, was also selected as an evaluation metric. It could reflect the current performance and learning ability of the agent. Another metric adopted was the intervention rate, which reflected the frequency of human intervention and guidance. The intervention rate can be represented in two ways, i.e., "count by episode" and "count by step". The former one was calculated based on the total number of steps guided by a human in a specific episodic interval, and the latter one can be calculated based on the number of episodes intervened by a human. The success rate was defined as the percentage of the successful episodes among total episodes through the testing process. The vehicle dynamics states, including the lateral acceleration and the yaw rate, were selected to evaluate dynamic performance and stability of the agent vehicle.

**Data availability**

The data that supports the findings of this study is available from the corresponding author upon reasonable request.

**Acknowledgments**

This work was supported in part by the SUG-NAP Grant of Nanyang Technological University, the A*STAR Grant (No. 1922500046), Singapore, and the Alibaba Group through Alibaba Innovative Research (AIR) Program and Alibaba-NTU Singapore Joint Research Institute (JRI) (No. AN-GC-2020-012).

**Author contributions**

J. Wu developed the algorithms, designed and performed experiments, processed and analyzed data, interpreted results, and wrote the paper. Z. Huang performed experiments, analyzed data, interpreted results, and wrote the paper. C. Huang, Z. Hu, P. Hang and Y. Xing designed and developed the experimental platforms, and analyzed data. C. Lv supervised the project, designed the methodology and experiments, analyzed data, interpreted results, and led the writing of the paper. All authors reviewed the manuscript.

**Competing interests**

The authors declare no competing interests.

Supplementary Materials for

# Human-in-the-Loop Deep Reinforcement Learning with Application to Autonomous Driving

**Authors**

Jingda Wu, Zhiyu Huang, Chao Huang, Zhongxu Hu, Peng Hang, Yang Xing, Chen Lv*

**Affiliation**

School of Mechanical and Aerospace Engineering, Nanyang Technological University, 50 Nanyang Ave, 639798, Singapore

*Corresponding author. Email: lyuchen@ntu.edu.sg

**Supplementary Note 1**

The architecture of the proposed Hug-DRL algorithm.

---
**Algorithm S1** Hug-DRL
---

Initialize the critic networks $Q_1$, $Q_2$, and actor network $\mu$ with random parameters.

Initialize the target networks with the same parameters as their counterparts. Initialize experience replay buffer $D$.

Initialize experience replay buffer $D$

**for** $epoch = 1$ **to** $M$ **do**

  **for** $t = 1$ **to** $T$ **do**

    **if** the human driver does not intervene

      let $I(s_t) = 0$, and select action with exploration noise $a_t \leftarrow a_t \sim \mu(s_t) + \epsilon$, with $\epsilon \sim \mathcal{N}(0, \sigma)$

    **Otherwise**

      let $I(s_t) = 1$, and adopt human action $a_t \leftarrow a_t^{human}$

    Observe reward $r_t$ and the new state $s_{t+1}$, terminal signal $d_t$, store transition tuple $\{s_t, a_t, r_t, d_t, s_{t+1}, I(s_t)\}$ in buffer $D$

    Sample minibatch of $N$ tuples from $D$, calculate the target value of critic network as $y_i \leftarrow r_i + \gamma(1 - d_i) \min_{j=1,2} Q_j'\left(s_{i+1}, \mu'(s_{i+1})\right)$, and update the critic networks by: $\theta^{Q_j} \leftarrow argmin_{\theta^{Q_j}} N^{-1} \sum_i^N \left(y_i - Q_j(s_i, a_i)\right)^2$

    **if** $t$ mod $d$ **then**

      Update the policy network $\mu$ by the proposed loss function:

      $\nabla_{\theta^\mu} L^\mu = N^{-1}\{(-\nabla_a Q_1(s,a)|_{s=s_t, a=\mu(s_t)} \nabla_{\Theta^\mu} \mu(s)|_{s=s_t}) + (\nabla_{\Theta^\mu}(\omega_I \cdot \|a - \mu(s)\|^2)|_{s=s_t, a=a_t}) \cdot I(s_t)\}$

      where $\omega_I = \lambda^k \cdot \{\max[\sup_{(s_t, a_t) \in \mathcal{D}} \exp(Q_1(s_t, a_t) - Q_1(s_t, \mu(s_t|\Theta^\mu))), 1] - 1\}$.

      Update the target networks:

      $\theta^{Q'} \leftarrow \tau \theta^Q + (1 - \tau)\theta^{Q'}$ for both target critic networks.

      $\theta^{\mu'} \leftarrow \tau \theta^\mu + (1 - \tau)\theta^{\mu'}$ for the target actor network.

    **end if**

  **end for**

**end for**

**Supplementary Note 2**

The Experimental Platform

    The experiment was carried out on a simulated driving platform that runs the CARLA simulator, which provides open-source codes supporting the flexible specification of sensor suites, environmental conditions, and full control of all static and dynamic automated-driving-related modules. Our platform consists of a computer equipped with an NVIDIA GTX 2080 Super GPU, three joint heads-up monitors, Logitech G29 steering wheel suit, and driver seat. During the training process, participants observed the frontal-view images fixed to the ego vehicle, from the screen while the DRL agent received the semantic images as state variables. Both the control frequency of the ego vehicle and the frequency of data recording are 20 Hz. After human-in-the-loop training, the generated mature strategy for autonomous driving could be leveraged to execute in various testing scenarios. In this study, all the control codes and algorithms were programmed in the Python environment, and the deep neural networks were built leveraging the Pytorch framework.

## Supplementary Note 3

The architecture of IA-RL algorithm.

---

**Algorithm S2** IA-RL (off-policy version)

---

Initialize the critic networks $Q_1$, $Q_2$, and actor network $\mu$ with random parameters.

Initialize the target networks with the same parameters as their counterparts. Initialize the experience replay buffer $D$.

Initialize experience replay buffer $D$

**for** $epoch = 1$ **to** $M$ **do**

  **for** $t = 1$ **to** $T$ **do**

    **if** the human participants does not intervene

      let $I(s_t) = 0$, and select action with exploration noise $a_t \leftarrow a_t \sim \mu(s_t) + \epsilon$, with $\epsilon \sim \mathcal{N}(0, \sigma)$

    **Otherwise**

      let $I(s_t) = 1$, and adopt human action $a_t \leftarrow a_t^{human}$

    Observe reward $r_t$ and the new state $s_{t+1}$, terminal signal $d_t$, store transition tuple $\{s_t, a_t, r_t, d_t, s_{t+1}, I(s_t)\}$ in buffer $D$

    Sample minibatch of $N$ tuples from $D$, calculate the target value of critic network as $y_i \leftarrow r_i + \gamma(1 - d_i) \min_{j=1,2} Q_j'\left(s_{i+1}, \mu'(s_{i+1})\right)$, and update critic networks by: $\theta^{Q_j} \leftarrow argmin_{\theta^{Q_j}} N^{-1} \sum_i^N \left(y_i - Q_j(s_i, a_i)\right)^2$

    **if** $t$ mod $d$ **then**

      Update policy network $\mu$ by the proposed loss function:

      $\nabla_{\theta^\mu} L^\mu = N^{-1} \sum_i^N \left\{ \left(-\nabla_a Q_1(s_i, a_i)|_{a=\mu(s_i)} \nabla_{\theta^\mu} \mu(s_i)\right) \cdot [1 - I(s_i)] + \left(\nabla_{\theta^\mu} \left(\omega_I \left(a_i - \mu(s_i)\right)^2\right)\right) \cdot I(s_i) \right\}$

      Update the target networks:

      $\theta^{Q'} \leftarrow \tau \theta^Q + (1 - \tau)\theta^{Q'}$ for both target critic networks.

      $\theta^{\mu'} \leftarrow \tau \theta^\mu + (1 - \tau)\theta^{\mu'}$ for the target actor network.

    **end if**

  **end for**

**end for**

## Supplementary Note 4

The architecture of HI-RL algorithm.

---

**Algorithm S3** HI-RL (off-policy version)

---

Initialize the critic networks $Q_1$, $Q_2$, and actor network $\mu$ with random parameters.

Initialize the target networks with the same parameters as their counterparts. Initialize experience replay buffer $D$.

Initialize the experience replay buffer $D$

**for** $epoch = 1$ **to** $M$ **do**

  **for** $t = 1$ **to** $T$ **do**

    **if** the human driver does not intervene

      let $I(s_t) = 0$, and select action with exploration noise $a_t \leftarrow a_t \sim \mu(s_t) + \epsilon$, with $\epsilon \sim \mathcal{N}(0, \sigma)$

    **Otherwise**

      let $I(s_t) = 1$, and adopt human action $a_t \leftarrow a_t^{human}$

    Observe reward $r_t$ and the new state $s_{t+1}$, terminal signal $d_t$, store transition tuple $\{s_t, a_t, r_t, d_t, s_{t+1}, I(s_t)\}$ in buffer $D$

    Sample minibatch of $N$ tuples from $D$, calculate the target value of critic network as $y_i \leftarrow r_i + \gamma(1 - d_i)\min_{j=1,2} Q'_j(s_{i+1}, \mu'(s_{i+1}))$ , and update critic networks by: $\theta^{Q_j} \leftarrow argmin_{\theta^{Q_j}} N^{-1} \sum_i^N \left(y_i - Q_j(s_i, a_i)\right)^2$

    **if** $t$ mod $d$ **then**

      Update the policy network $\mu$ by the loss function:

      $\nabla_{\theta^\mu} L^\mu = N^{-1} \sum_i^N \left\{\left(-\nabla_a Q_1(s_i, a_i)|_{a=\mu(s_i)} \nabla_{\theta^\mu} \mu(s_i)\right)\right\}$

      Update the target networks:

      $\theta^{Q'} \leftarrow \tau\theta^Q + (1 - \tau)\theta^{Q'}$ for both target critic networks.

      $\theta^{\mu'} \leftarrow \tau\theta^\mu + (1 - \tau)\theta^{\mu'}$ for the target actor network.

    **end if**

  **end for**

**end for**

## Supplementary Note 5

The architecture of Vanilla-DRL algorithm.

**Algorithm S4** Vanilla-DRL (Twin delayed deep deterministic policy gradient)

Initialize the critic networks $Q_1$, $Q_2$, and actor network $\mu$ with random parameters.

Initialize the target networks with the same parameters as their counterparts. Initialize the experience replay buffer $D$.

**for** $epoch = 1$ **to** $M$ **do**

    **for** $t = 1$ **to** $T$ **do**

        select action with exploration noise $a_t \leftarrow a_t \sim \mu(s_t) + \epsilon$, with $\epsilon \sim \mathcal{N}(0, \sigma)$

        Observe reward $r_t$ and the new state $s_{t+1}$, terminal signal $d_t$, store transition tuple $\{s_t, a_t, r_t, d_t, s_{t+1}\}$ in buffer $D$

        Sample minibatch of $N$ tuples from $D$, calculate the target value of critic network as $y_i \leftarrow r_i + \gamma(1 - d_i) \min_{j=1,2} Q'_j(s_{i+1}, \mu'(s_{i+1}))$, and update critic networks by: $\theta^{Q_j} \leftarrow \arg\min_{\theta^{Q_j}} N^{-1} \sum_i^N \left( y_i - Q_j(s_i, a_i) \right)^2$

        **if** $t \bmod d$ **then**

            Update the policy network $\mu$ by the loss function:

            $\nabla_{\theta^\mu} L^\mu = N^{-1} \sum_i^N \left\{ \left( -\nabla_a Q_1(s_i, a_i) |_{a=\mu(s_i)} \nabla_{\theta^\mu} \mu(s_i) \right) \right\}$

            Update the target networks:

            $\theta^{Q'} \leftarrow \tau \theta^Q + (1 - \tau) \theta^{Q'}$ for both target critic networks.

            $\theta^{\mu'} \leftarrow \tau \theta^\mu + (1 - \tau) \theta^{\mu'}$ for the target actor network.

        **end if**

    **end for**

**end for**

**Supplementary Note 6**

Data collection and model training of the Vanilla imitation learning-based strategy

The data collection session required human participants to complete driving tasks in scenario 1, and the state inputs (frontal-view images) and action outputs (steering angles) were recorded throughout the demonstration process. To solve the distributional shift problem, the data augmentation with added noise was adopted, referring to Supplementary Fig. 7b. We added random Gaussian noise into the steering command. Therefore, the human drivers would need to adjust their control actions to avoid unexpected risks, adapting to the varying states. In this way, more scenes and demonstrated datasets covering the situations recovered from potential failures were collected for the Vanilla-IL-based policy. Note that output action labels were still human steering actions without the added noise. Subfigures [i] and [iii] of Supplementary Fig. 7c illustrate the effect of noise-added data augmentation.

For offline training, the state or scene representation was the semantic image captured from the ego-vehicle with a size of 45×80 pixels. The policy network was a convolution neural network (CNN) that took the semantic images as input and output the steering controls. We constantly sampled a batch of state-action pairs from the dataset to adjust the parameters of the network. The loss function can be expressed as:

$$L^C(\theta^C) = \mathop{\mathbb{E}}_{(s_t, y_t) \sim D_{IL}} [(C(s_t) - y_t)^2] \qquad \text{(Eq. S1)}$$

where $C$ denotes the CNN-based imitation network, $\theta^C$ the parameters of the network, $s_t$ is the matrix of input data at the timestep $t$, $y_t$ is the labeled action of the human participant, and $D_{IL}$ is the dataset. The detailed parameters are listed in Supplementary Table 8 and 10.

Data collection and model training of the DAgger strategy

The data collection session requires human participants to complete one episode in scenario 1 with the same input/output as that of the Vanilla-IL method. And the human demonstration data will be utilized to pre-train the DAgger. To solve the distributional shift problem, the DAgger method allows the pre-trained agent to perform explorations in multiple episodes. When the distribution of the training data shifted to that of untrained situations, the human participants were required to intervene and share the control authority with the agent, aiming at correcting the ego vehicle's undesirable behaviours. The control authority assignment mechanism can be given by:

$$a_t = \beta a_t^{agent} + (1 - \beta) a_t^{human} \qquad \text{(Eq. S2)}$$

Despite the shared control, the DAgger imitation learning would only record the actions from human participants into the dataset and learn from those recorded demonstration data. The loss function can be expressed as:

$$L^D(\theta^D) = \mathop{\mathbb{E}}_{(s_t, y_t) \sim D_{DAgger}} [(C(s_t) - y_t)^2] \qquad \text{(Eq. S3)}$$

where $D$ denotes the CNN-based DAgger imitation network, $\theta^D$ refers to the parameters of the network, $s_t$ is the matrix of input data at the timestep $t$, $y_t$ is the labeled action of the human participant, and $D_{DAgger}$ is the dataset. The schematic diagram of the DAgger method is provided in Supplementary Fig. 8, and the detailed parameters are listed in Supplementary Tables 9 and 10.

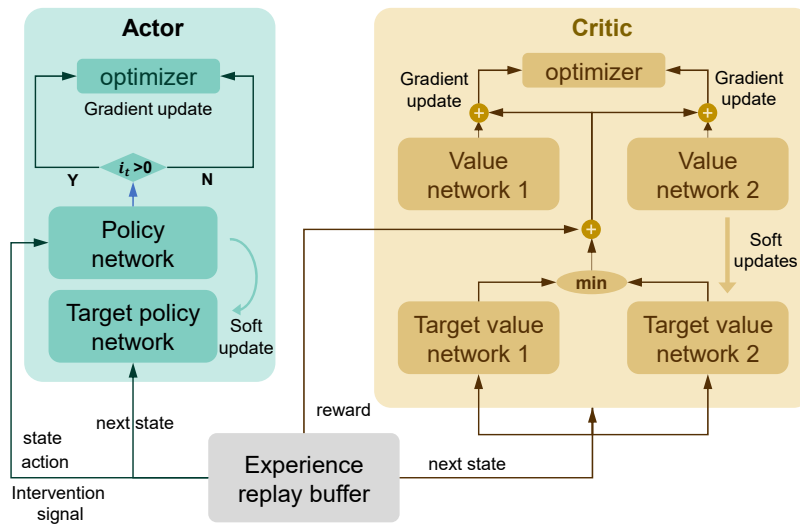

**Supplementary Fig. 1. The architecture of the adopted deep reinforcement learning algorithm, i.e., the twin delayed deep deterministic policy gradient (TD3) algorithm.** Within the actor-critic principle, the policy network conducts the control task based on the state input, and the value networks generate evaluations that help to optimize the policy network.

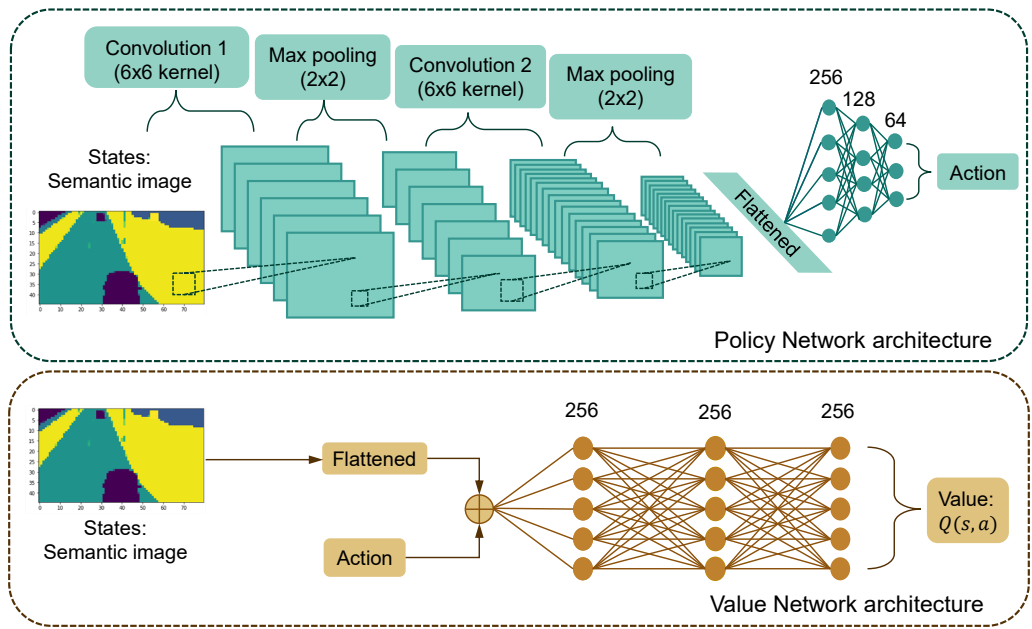

**Supplementary Fig. 2. The structure of the policy and value networks of deep reinforcement learning.** The semantic images are fed to both the policy and value networks. The policy network processes images with twice convolution-pooling operations and then flattens the data and sends the extracted features to the fully connected layers, which eventually outputs the control action.

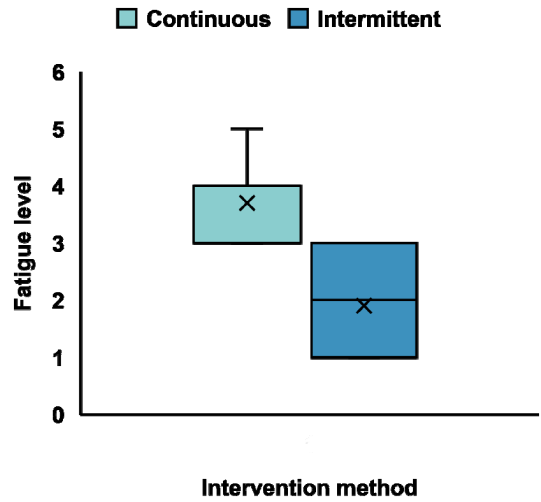

**Supplementary Fig. 3. Evaluation of the subjective responses to the question on workload during experiments.** The human workload levels under the continuous and intermittent guidance modes were rated at 3.70 ± 0.67, and 1.90 ± 0.88, respectively (rating scale from 1: very low, to 3: normal, to 5: very high). A significant difference between the two groups was found (p=6.73e-5). The utilized data of this figure was listed in Supplementary Table 4.

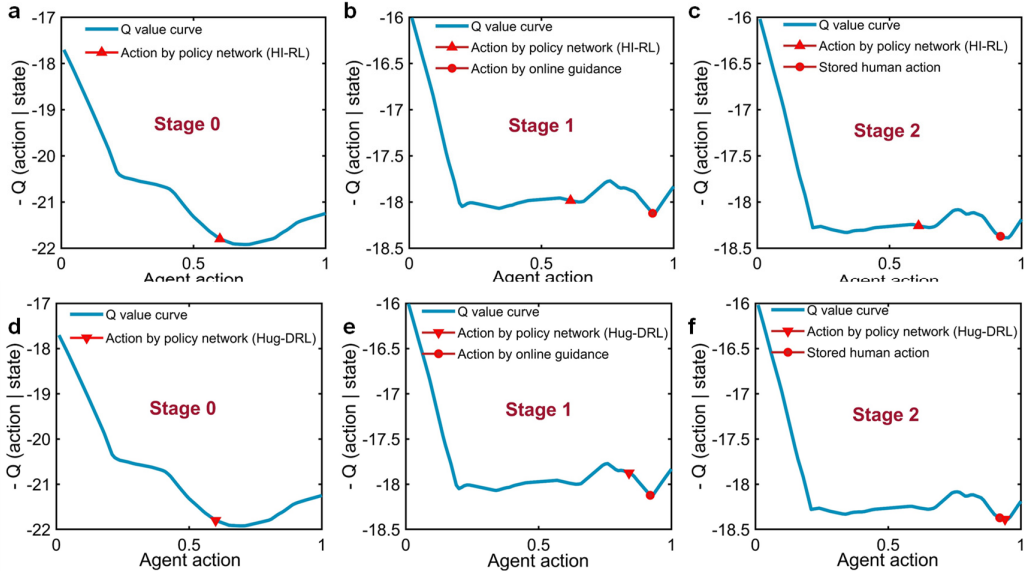

**Supplementary Fig. 4. Illustration of the finetuning stage of the Hug-DRL and HI-RL methods.** The data was collected from one timestep of a typical participant's experimental results, where the input state and human guidance action were kept the same in two methods for comparison. The horizontal axis represents all possible actions of the agent, which is the steering wheel angle in this work. [0, 1] corresponds to the entire range of the steering wheel angle from the extreme left position to the extreme right position. **a,** Action evaluation by the value network of the pre-trained HI-RL agent at stage 0 (before fine-tuning). At stage 0: before fine-tuning, the actions given by the policy network of HI-RL agent were away from the optimal solution evaluated by the value network. **b,** Action evaluation by the value network of HI-RL during stage 1 (the fine-tuning stage). The evaluation curve of the value network was updated based on the guidance action provided by the human participant, resulting in a new global optimal solution. Yet, the update of the HI-RL's policy network remained unremarkable. **c,** Action evaluation by the value network of the HI-RL during stage 2 (the post-finetuning stage). After the fine-tuning, the policy network of the HI-RL failed to reach the global optimal point, as the agent fell into the local optima trap resulted from the gradient-descent update principle. **d,** Action evaluation by the Hug-DRL agent at stage 0. The actions given by the policy network were also away from the optimal solution evaluated by the value network. **e,** Action evaluation by the Hug-DRL at stage 1. The evaluation curve of the value network was updated based on the guidance action provided by the human participant and generated a new global optimal solution similar to the situation of HI-RL. The policy network of the Hug-DRL was able to approach the new optima thanks to the redesigned policy function in the proposed method. **f,** Action evaluation by the value network of the Hug-DRL at stage 2. The policy network of the Hug-DRL maintained a good performance and avoided the local optima trap during the post-fine-tuning stage.

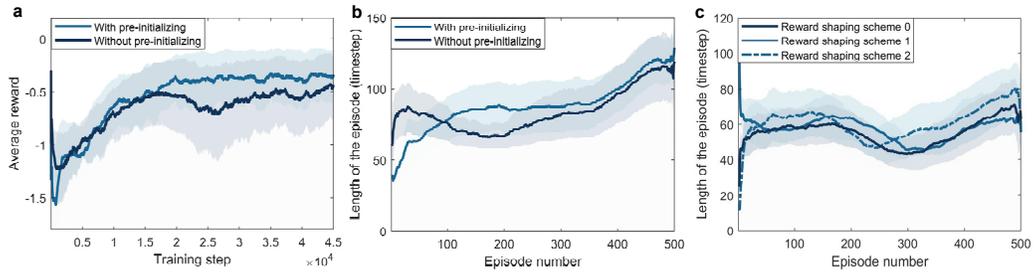

**Supplementary Fig. 5. Ablation study of the pre-initialization and reward shaping. a,** The averaged reward of Hug-DRL during the training session obtained with/without the pre-initialization. **b,** The length of the episode (counted by time step of the simulator) of Hug-DRL during the training session obtained with/without the pre-initialization. **c,** The length of the episode of Hug-DRL during the training session obtained with different reward-shaping schemes. Scheme 0 denotes no reward-shaping scheme, Scheme 1 and 2 represent the two approaches described in the Method Section of the main text, respectively.

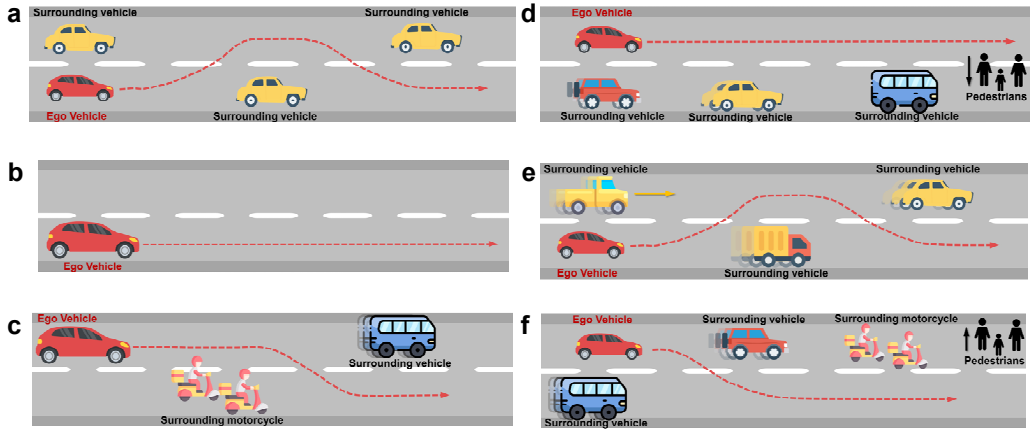

**Supplementary Fig. 6 Schematic diagram of the scenarios for training and testing of the autonomous driving agent. a,** Scenario 0 serves as a simple situation with all surrounding vehicles being set as stationary states. It was utilized only for the training stage. Besides, two pedestrians were spawned at random positions in some episodes, but were not shown in the figure due to the unfixability. **b,** Scenario 1 was used to test the steady driving performance of the agent on the freeway, with the removal of all surrounding traffic participants. It was used to evaluate the anti-overfitting performance of the generated driving policy. **c-f,** Scenarios 2 to 5 were used to test the adaptiveness of the obtained policy in unseen situations shielded from the training stage. The moving pedestrians, motorcycles, and buses were added into the traffic scenarios. Since the interactive relationships between the ego vehicle and the traffic participants were changed, the expected trajectories of the ego vehicle should be different from those in the training process. These were set to evaluate the scene understanding ability and its adaptiveness and robustness of the autonomous driving agent.

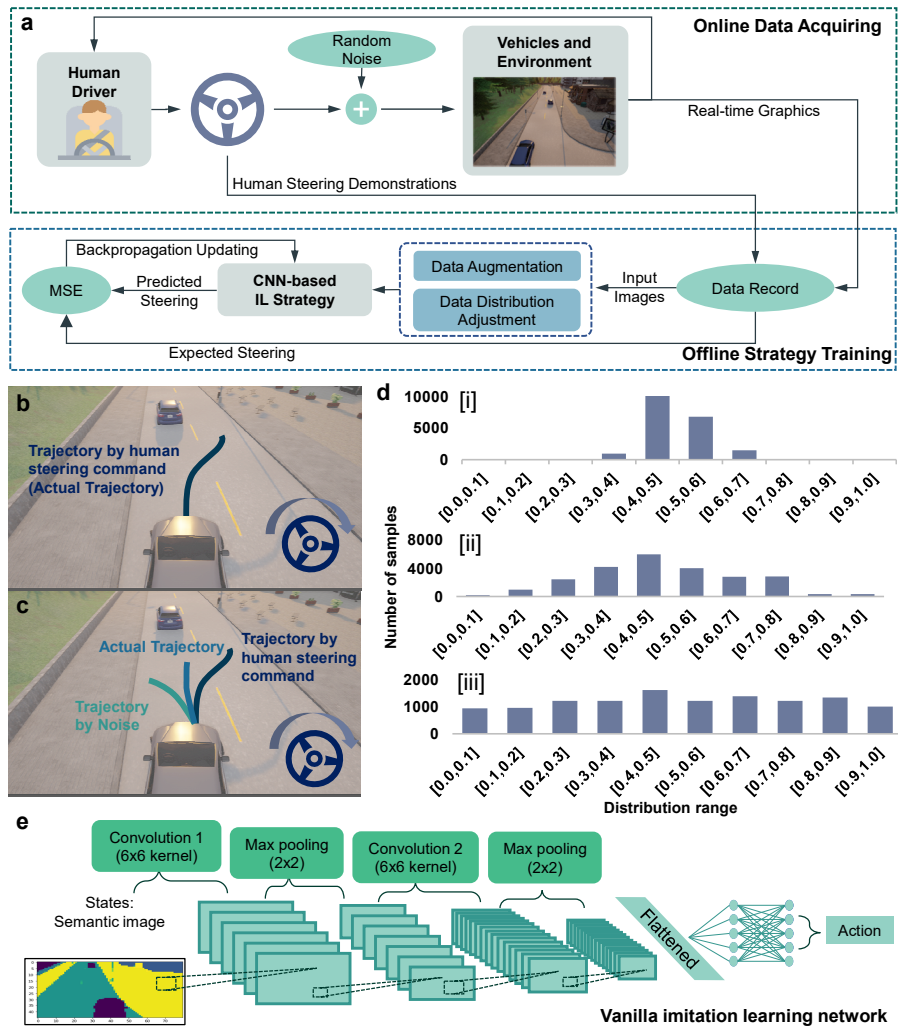

**Supplementary Fig. 7. Implementation details of the Vanilla imitation learning-based strategy for autonomous driving. a,** The architecture of the Vanilla-IL-based method. Human participants provide real-time control input through the steering wheel as demonstrations in the simulated driving environment, where corresponding images and control commands were recorded for subsequent training. The obtained data were firstly processed with augmentation and adjustment and were subsequently used for the training of the convolutional neural network. The Gaussian noise signal was injected into the original output of the steering wheel angle to generate many demonstration scenes. **b,** The schematic diagram of the conventional data sampling of the human participant. **c,** the schematic diagram of the augmented data sampling of the human participant under injected noise. Note that the recorded actions were only human actions, while the noise was filtered in the samples. **d,** The performance of demonstrated data augmentation and related adjustment. Subfigure [i] shows the human demonstrated action sets without added noise; subfigure [ii] shows the action distribution after noise-based augmentation, and subfigure [iii] is the distribution histogram further processed by data augmentation and adjustment, which is the adopted training data in the imitation learning. **e,** The architecture and parameters of the convolutional neural network in the Vanilla-IL-based method, where input variables are kept the same as those in the DRL for comparison.

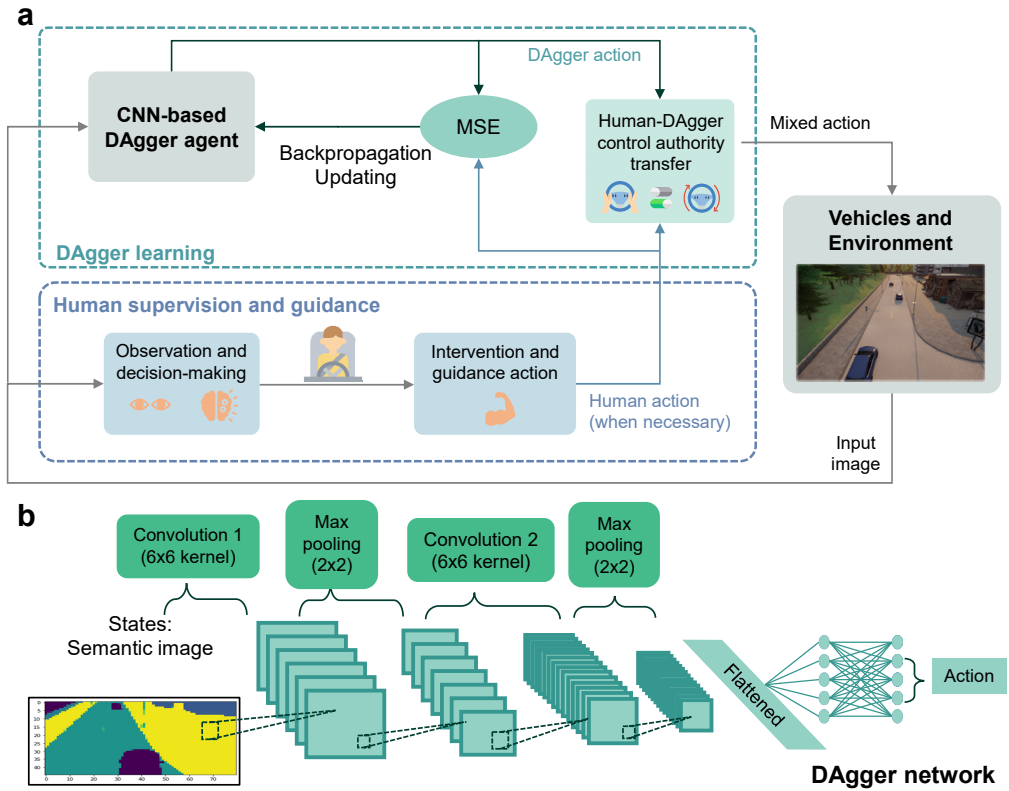

**Supplementary Fig. 8. Implementation details of the DAgger imitation learning-based strategy for autonomous driving. a,** The architecture of the DAgger IL-based method. Human participants perform driving control behaviors at the initial stage whereas the DAgger agent learns from the demonstration data. The control authority of the agent during training was mostly dominated by the DAgger. The human participants were required to intervene and adjust those risky actions given by the DAgger especially when the distributional shift problems occur. The human demonstration data would be utilized for the further training of the DAgger agent by minimizing the loss between the DAgger actions and human actions. **b,** The architecture and parameters of the convolutional neural network in the DAgger-based method. The input variables are kept the same as those in the DRL for comparison.

**Supplementary Table 1**
Hyperparameters used in the DRL algorithms. These parameters were universally applied to all involved DRL algorithms.

| Parameter | Description (unit) | Value |
| --- | --- | --- |
| *Replay buffer size* | The capacity of the experience replay buffer | 384000 |
| *Maximum steps* | Cutoff step number of the "cold-start" training process | 50000 |
| *Minibatch size (N)* | Capacity of minibatch | 128 |
| *Learning rate of actors* | Initial learning rate (policy/actor networks) with Adam optimizer | 0.0005 |
| *Learning rate of critics* | Initial learning rate (value/critic networks) with Adam optimizer | 0.0002 |
| *Initialization* | Initialization method of Dense layers of the network | *he_initializer* |
| *Activation* | Activation method of layers of the network | relu |
| *Initial exploration* | Initial exploration rate of noise in $\epsilon - greedy$ | 1 |
| *Final exploration* | Cutoff exploration rate of noise in $\epsilon - greedy$ | 0.01 |
| *Gamma ($\gamma$)* | Discount factor of the action-value function of value network | 0.95 |
| *Soft updating factor* | Parameter transferring speed from policy/value networks to target policy/value networks | 0.001 |
| *Noise scale* | Noise amplitude of action in TD3 algorithm | 0.2 |
| *Policy delay (d)* | Updating frequency of value networks over policy networks | 2 |

**Supplementary Table 2**
ANOVA analysis for the training reward of different DRL methods with the "cold-start" condition. Significant values at the α = 0.05 level are marked bold.

| Metric 1: ANOVA Table | | | | | |
|---|---|---|---|---|---|
| Source | Sum Sq. | d.f. | Mean Sq. | $F$-value | $p$-value |
| between-groups variation | 0.07 | 3 | 0.02 | 27.99 | **1.61e-9** |
| within-groups variation | 0.03 | 36 | 0.00087 | | |
| Total | 0.10 | 39 | | | |
| Metric 2: Multiple comparisons between groups | | | | | |
| Compared pair | Difference between estimated group means | | | $p$-value | |
| Proposed Hug-DRL vs. IA-RL | 0.06 | | | **6.57e-5** | |
| Proposed Hug-DRL vs. HI-RL | 0.10 | | | **7.49e-8** | |
| Proposed Hug-DRL vs. Vanilla-DRL | 0.11 | | | **2.84e-6** | |

**Supplementary Table 3**
ANOVA analysis for the averaged length in timestep of different DRL methods with the "cold-start" condition. Significant values at the α = 0.05 level are marked bold.

| Metric 1: ANOVA Table | | | | | |
|---|---|---|---|---|---|
| Source | Sum Sq. | d.f. | Mean Sq. | $F$-value | $p$-value |
| between-groups variation | 2541.99 | 3 | 847.33 | 14.04 | **3.22e-6** |
| within-groups variation | 2172.69 | 36 | 60.35 | | |
| Total | 4714.68 | 39 | | | |
| Metric 2: Multiple comparisons between groups | | | | | |
| Compared pair | Difference between estimated group means | | | $p$-value | |
| Proposed Hug-DRL vs. HI-RL | 3.63 | | | 0.22 | |
| Proposed Hug-DRL vs. IA-RL | 7.11 | | | **0.04** | |
| Proposed Hug-DRL vs. Vanilla-DRL | 21.05 | | | **8.10e-6** | |

**Supplementary Table 4**

The subjective evaluation scores of participants under the two guidance modes. Significant value at the α = 0.05 level is marked bold.

| Participant information | | | Subjective scores regarding workload (1: very low, 5: very high) | |
|---|---|---|---|---|
| Index | Sex | Age | Continuous mode | Intermittent mode |
| 1 | Male | 26 | 3 | 1 |
| 2 | Male | 32 | 5 | 3 |
| 3 | Male | 28 | 4 | 3 |
| 4 | Male | 24 | 3 | 2 |
| 5 | Male | 29 | 4 | 2 |
| 6 | Male | 27 | 3 | 1 |
| 7 | Male | 33 | 3 | 1 |
| 8 | Male | 30 | 4 | 3 |
| 9 | Female | 33 | 4 | 2 |
| 10 | Female | 30 | 4 | 1 |
| | | | | |
| Mean | | | 3.7 | 1.9 |
| S.D. | | | 0.67 | 0.88 |
| *p*-value of paired t-test between the two modes | | | **6.74e-5** | |

**Supplementary Table 5**

Kruskal-Wallis ANOVA analysis for the training rewards obtained in the non-guided sections during the "cold-start" training by the proposed Hug-DRL method with proficient and non-proficient participants. The standard DRL approach was taken as the baseline for comparison. Significant values at the α = 0.05 level are marked bold.

| Metric 1: ANOVA Table | | | | | |
|---|---|---|---|---|---|
| Source | Sum Sq. | d.f. | Mean Sq. | $F$-value | $p$-value |
| Between-groups variation | 1680.00 | 2 | 840.00 | 21.68 | **1.96e-5** |
| Within-groups variation | 567.50 | 27 | 21.02 | | |
| Total | 2247.50 | 29 | | | |
| Metric 2: Multiple comparisons between groups | | | | | |
| Compared pair | Difference between estimated group means | | | $p$-value | |
| Proficient Hug-DRL vs. Non-proficient Hug-DRL | 6 | | | 0.28 | |
| Proficient Hug-DRL vs. Standard DRL | 18 | | | **1.44e-5** | |
| Non-proficient Hug-DRL vs. Standard DRL | 12 | | | **0.0065** | |

**Supplementary Table 6**

Kruskal-Wallis ANOVA analysis for the overall training reward obtained in the "cold-start" training by the proposed Hug-DRL method with proficient and non-proficient participants. The standard DRL approach was taken as the baseline for comparison. Significant values at the α = 0.05 level are denoted with boldface type.

| Metric 1: ANOVA Table | | | | | |
|---|---|---|---|---|---|
| Source | Sum Sq. | d.f. | Mean Sq. | $F$-value | $p$-value |
| Between-groups variation | 1804.2 | 2 | 902.1 | 23.28 | **8.80e-6** |
| Within-groups variation | 443.3 | 27 | 16.42 | | |
| Total | 2247.5 | 29 | | | |
| Metric 2: Multiple comparisons between groups | | | | | |
| Compared pair | Difference between estimated group means | | | $p$-value | |
| Proficient Hug-DRL vs. Non-proficient Hug-DRL | 7.8 | | | 0.11 | |
| Proficient Hug-DRL vs. Standard DRL | 18.9 | | | **4.72e-6** | |
| Non-proficient Hug-DRL vs. Standard DRL | 11.1 | | | **0.013** | |

**Supplementary Table 7**

Configuration of the experimental platform.

| | | |
|---|---|---|
| Driving platform | Simulation rendering software | *CARLA* |
| | Steering wheel suit | *Logitech G29* |
| | CPU of the host computer | *Intel i9-9900k* |
| | GPU of the host computer | *NVIDIA GTX2080 Super* |
| | Monitoring device | *Joint heads-up monitors ×3* |
| | Other equipment | *Driver seat suit* |
| Simulation configuration | Control frequency (train) | *20Hz* |
| | Control (test) | *50Hz* |
| | Spawned vehicle type | *Sedan* |
| | Programming script | *Python* |
| | Neural network toolbox | *Pytorch* |

**Supplementary Table 8**

Network details

The details of the reinforcement learning architecture, applied to all related DRL algorithms

| Parameter | Value |
|---|---|
| Input Image shape | [80,45,1] |
| Policy/actor Network Convolution Filter Features | [6,16] (kernel size 6×6) |
| Policy/actor Network Pooling Features | Maxpooling (Stride 2) |
| Policy/actor Network Fully Connected Layer Features | [256,128,64] |
| Value/critic Network Fully Connected Layer Features | [256,256,256] |

The details of the imitation learning architecture, applied to Vanilla-IL and DAgger algorithms

| Parameter | Value |
|---|---|
| Input Image shape | [80,45,1] |
| Network Convolution Filter Features | [6,16] (kernel size 6×6) |
| Network Pooling Features | Maxpooling (Stride 2) |
| Network Fully Connected Layer Features | [256,128,64] |

The architecture of the NGU network, applied to Reward shaping scheme 2

| Parameter | Value |
|---|---|
| Input Image shape | [80,45,1] |
| Network Convolution Filter Features | [6,16] (kernel size 6×6) |
| Network Pooling Features | Maxpooling (Stride 2) |
| Network Fully Connected Layer Features | [128] |
| Network Initial Learning rate | 0.0001 |

**Supplementary Table 9**

Hyperparameters used in the Vanilla imitation learning algorithm.

| Parameter | Description (unit) | Value |
|---|---|---|
| *Learning rate* | Initial learning rate with Adam optimizer | 0.00001 |
| *Initialization* | Initialization method of Dense layers of the network | *he_initializer* |
| *Activation* | Activation method of layers of the network | relu |
| *Noise scale* | Noise amplitude of action in data augmentation | 0.1 |

**Supplementary Table 10**

Hyperparameters used in the DAgger algorithm.

| Parameter | Description (unit) | Value |
|---|---|---|
| *Learning rate* | Initial learning rate with Adam optimizer | 0.00001 |
| *Initialization* | Initialization method of Dense layers of the network | *he_initializer* |
| *Activation* | Activation method of layers of the network | relu |
| *Maximum episodes* | Cutoff episode number of the training process | 50 |
| *Batch size* | Capacity of minibatch | 128 |

**Supplementary Video 1**

The animation of the ego vehicle trajectories throughout the training process. The curves refer to the varying trajectories of the ego vehicle. The upper plot shows the trajectories of the ego vehicle under the proposed Hug-DRL method during the training process, and the lower one corresponds to the trajectories of the ego vehicle under the baseline B, i.e., the standard DRL TD3 method. The gray blocks and black solid lines represent the obstacle vehicles and road boundaries, respectively.

The video can be accessed from the link below:

https://drive.google.com/file/d/1oCl3BnBiCDqoRWoeLkKDnP_H3GMFzPxk/view?usp=sharing